\documentclass[10pt,journal,cspaper,compsoc]{IEEEtran}
\usepackage{epsfig,graphicx}
\usepackage{xcolor}
\usepackage{amsfonts,amsmath,amssymb}
\usepackage{algorithm}
\usepackage{algpseudocode}
\def\by{{\mathbf y}}
\def\bx{{\mathbf x}}
\def\bv{{\mathbf v}}
\def\bu{{\mathbf u}}
\def\bc{{\mathbf c}}
\def\be{{\mathbf e}}
\def\cS{{\mathcal S}}
\def\diag{\text{{diag}}}
\def\tr{\text{{tr}}}

\ifCLASSOPTIONcompsoc
\else
\fi
\ifCLASSINFOpdf
\else
\fi
\begin{document}
%
\title{Fast Greedy Algorithm\\  for Subspace Clustering\\ from Corrupted and Incomplete Data}
%
%
%
%

\author{Alexander~Petukhov, Inna~Kozlov%
\IEEEcompsocitemizethanks{\IEEEcompsocthanksitem A.Petukhov is with the Department
of Mathematics, University of Georgia, Athens,
GA, 30602.\protect\\
E-mail: see http://www.math.uga.edu/\~{}petukhov
\IEEEcompsocthanksitem I.Kozlov is CTO, Algosoft-Tech USA.}
\thanks{}}

\IEEEcompsoctitleabstractindextext{%
\begin{abstract}
We describe the Fast Greedy Sparse Subspace Clustering (FGSSC)  algorithm providing an efficient method 
for clustering data belonging to a few low-dimensional linear or affine subspaces. The main difference of our algorithm from predecessors is its ability to work with noisy data having a high rate of  erasures (missed entries with the known coordinates) and errors (corrupted entries with  unknown coordinates). 
\par
The   greedy approach consists in usage of a basic algorithm on data out of the~domain of the basic algorithm reliability. When the algorithm fails and its output looses some desirable features of the solution, usually it still brings some information about "the largest features" of the solution. The greedy algorithm extracts those  features and uses them in an iterative way by launching the basic algorithm with the additional information mined out on the previous iterations. Such scheme requires an additional time consuming  algorithm loop. We discuss here how to implement the fast version of the greedy algorithm with the maximum efficiency whose greedy strategy is incorporated into  iterations of the basic algorithm.
\par
We provide numerical evidences that,  in 
the subspace clustering capability,  the fast greedy algorithm outperforms not only  the existing state-of-the art SSC algorithm taken by the authors as a basic algorithm but also the recent GSSC algorithm. At the same time, its computational cost is only slightly higher than the cost of SSC.
\par
The numerical evidence of the algorithm significant  advantage is presented for a few synthetic models as well as for the Extended Yale B dataset of facial images. In particular, the face recognition misclassification rate turned out to be 6--20 times lower than for the SSC algorithm.  We provide also the numerical evidence that the FGSSC  algorithm is  able to perform clustering of corrupted data efficiently  even when the sum of subspace  dimensions significantly exceeds the dimension of the ambient space.
\end{abstract}

\begin{keywords}
Subspace Clustering, Sparse Representations, Greedy Algorithm, Law-Rank Matrix Completion, Compressed Sensing, Face Recognition
\end{keywords}}
\maketitle

\IEEEdisplaynotcompsoctitleabstractindextext

%
\IEEEpeerreviewmaketitle

\section{Introduction}
We consider a greedy strategy  based algorithm for preprocessing on vector database 
necessary for subspace clustering. The problem of subspace clustering consists in classification of 
the vector data belonging to a few linear or affine low-dimensional subspaces of the high dimensional 
ambient space when neither subspaces or even their dimensions are not known, i.e., they have to be identified  from the same database. No dataset for algorithm learning is provided.
\par
The problem has a long history and always was considered as difficult. In spite of its closeness to the problem of linear regression, the presence of multiple subspaces brings combinatorial non-polynomial complexity. Additional hardness of the  settings considered in this paper is due to the presence of combined artifacts consisting of noise in the data, the errors at unknown locations. Moreover, a large data fraction may be missed. 
\par
There are many applied problems expecting the progress in subspace clustering algorithm. Among them the problems 
related to processing of visual information are especially popular. Typical problems are sorting databases consisting of images taken from a finite set of objects like faces, characters, symbols.
The special class of problems is constituted by motion segmentation problems which can be used for
better motion estimation, video segmentation, and video 2D to 3D conversion. The last successes 
in those areas are connected with the recent Sparse Subspace Clustering algorithm developed in
\cite{EV0}, \cite{EV} and thoroughly studied in \cite{SC}, \cite{SEC}.
\par
Using more formal definition, we have $N$ vectors $\{\by_l\}_{l=1}^N$ in $K$ linear or affine subspaces $\cS:=\{\cS^l\}_{l=1}^K$ with the dimensions $\{d_l\}_{l=1}^K$ of the $D$-dimensional Euclidean space $\mathbb R^D$.  We do not assume that those spaces do not have non-trivial intersections. However, we do assume that any one of those spaces is not a subspace of other one.
At the same time, the situation when one subspace is a subspace of a  sum of two or more subspaces from $\cS$  is allowed. Such settings  inspire the hope that when $N$ is significantly large  and the points are randomly and independently distributed on those planes some sophisticated algorithm can identify
those planes (subspaces) and classify the belongingness of each point to the found subspaces. 
Then the problem consists in finding a permutation matrix $\Gamma$ such that 
$$
[Y_1, \dots, Y_K]=Y\Gamma,
$$
where $Y\in \mathbb R^{D\times N}$ is an input matrix whose columns are the given points in an arbitrary random order, whereas  in $[Y_1, \dots, Y_K]$ is rearrangement of the matrix $Y$ in the accordance with the affiliation of the vectors with the subspaces $\cS^k$. 
\par
Let us discuss restrictions on the input data allowing expect some tractable output of the "ideal" clustering algorithm provided that such algorithm exists. Any space $\cS^k$ defined by the cluster $Y_k$ 
has dimension at most $\# Y_k-1$ for the linear space settings and at most $\# Y_k-2$ for the affine space settings. In addition to the dimension restriction, we  require that any cluster $Y_k$ cannot be split into 2 or more clusters satisfying the dimension condition.
\par
Of course, some of points may belong to the intersection of two or more subspaces, then such point may be assigned to one of those subspaces or to all of them. However, it is reasonable to assume that, with  probability 1, the points of $Y$ belong to only one of subspaces from $\cS$.
\par
Summarizing discussion above, we formulate the precise model of the problem/algorithm input and 
what is an expected algorithm outcome. 
\par
We start with the input model. The problem "generator" (synthetic or natural) consists of 2 parts: generation of subspaces and generation of data on those subspaces. 
\par
While the subspaces from $\cS$ can be generated randomly, they usually reflect the nature of the data and, in most cases,
they can be described deterministically. The data in the matrix $Y$ are formed as consecutive $N$ times random selection of a subspace $\cS^k$ and then the random point belonging to it.  We assume that random generator has distribution having probability 0 for any proper  linear (affine) subspace of  $\cS^k$. This property is instantly imply that almost for sure any selected vector of $\cS^k$ is a linear combination of other selected vectors from $\cS^k$ provided that the cardinality of that set exceeds  $d_k$ by at least 1 (or 2).
\par
What is expected as an ideal algorithm outcome? If the problem generator provided us with a sufficient number 
of points belonging to the given space $\cS^k$, we have right to expect that the algorithm will find
the cluster $Y_k$ including all its points, when $\# Y_k\ge d_k+1\, (\ge d_k+2)$. Since  the problem generator creates a finite number of points $N$, the probability that some of subspaces
are not represented by a significant number of points is not zero. No algorithm is able to identify such subspace for sure. So we have to assign those points as {\it outliers}. We also may think about outliers as points generated randomly from entire space $\mathbb R^D$. It is convenient to put all outliers into additional set $Y_0$ not corresponding to any of spaces from $\cS$ and to agree that all 
spaces in $\cS^k$ have enough representatives in $\{Y_k\}_{k=1}^K$.
\par
The problem of finding clusters $\{Y_k\}$ corresponding to linear/affine subspaces  is solved usually by means of finding the clusters in the similarity graph whose edges (to be more precise the weights of edges) characterize the interconnection between pairs of vertexes. In our  case, the popular method of clustering consists in making the points to play role of the vertexes, while the weights are  set from the coefficients of decomposition of the vectors through other vectors from the same space $\cS^k$. This idea looks as vicious circle. We are trying to identify the space $\cS^k$ accommodating the vector $\by_i$, using its linear decomposition  in the~remaining vectors from the (not found yet) cluster $Y_k\subset\cS^k$. However, the situation is not hopeless at all. In   \cite{EV}, the excellent suggestion to reduce  the decomposition problem formulated above  to solving the non-convex problem 
\begin{equation}
\label{CS}
\|\bc_i\|_0\to\min,\text{ subject to }\by_i=Y\bc_i,\, c_{i,i}=0, \,i=1,\dots,N;
\end{equation}
where $\|\bx\|_0$  is the Hamming weight of the vector $\bx$, $\|\bx\|_0:=\#\{x_j\ne0\}$, was presented. The problem of finding the sparsest solutions to (\ref{CS}), so-called, {\it Compressed Sensing} or {\it Compressive Sampling} (CS), underwent thorough study originated in \cite{CT}, \cite{D}, \cite{RV} and continued in hundreds of theoretical and applied papers.
\par
We have to emphasize that some requirements for the matrix $Y$ as well as topics of interest typical for CS settings are absolutely irrelevant to the problem above and
 may look too restrictive. Among those topics we mention the problem of the uniqueness of the solution to (\ref{CS}). Of course, the uniqueness of the solution is not important for us.
Moreover, the minimization of the Hamming weight is difficult and may look unnecessary in our case.
Indeed, if the sum of the spaces $\cS^k$ is  direct, any decomposition  fits our request. In this case, we can take, for example, the least $\ell^2$-norm solutions. When the sum is not direct  the minimization of the Hamming weight becomes very helpful but still unnecessary.
\par
In the ideal world of perfect computational precision and unlimited computational power, with probability one, the solutions 
of (\ref{CS}) would point out the elements of the appropriate $\cS^l$ by their non-zero decomposition coefficients. Provided that no vectors of wrong subspaces participate in decomposition of each column of $Y$, the matrix $C$ whose columns are $\bc_i$ allows perfectly reconstruct the structure of the subspaces in polynomial time. There are two obstacles on that way. 
\par
First, the precision of the input matrix $Y$ is usually not perfect. So the decompositions may pick up wrong vectors even if we are able to solve problem (\ref{CS}).  In this case, the problem of subspace clustering is considered for the similarity graph defined by the symmetric matrix $W:=|C|+|C^T|$. While, generally speaking, this problem cannot be solved in polynomial time, there exist practical algorithms allowing right clustering when the number 
of "false" interconnections of elements from different subspaces are not very dense and not very intensive. Following \cite{EV}, we will use some modification of  the spectral clustering algorithm from \cite{NWJ} which is specified in \cite{L} as "graph's random walk Laplacian".
\par
The second obstacle consists in non-polynomial complexity of problem (\ref{CS}) itself. The elegant solution allowing to overcome this obstacle is replacement of  non-convex problem  (\ref{CS}) with the convex problem
\begin{equation}
\label{CS1}
\|\bc_i\|_1\to\min,\text{ subject to }\by_i=Y\bc_i,\, c_{i,i}=0, \,i=1,\dots,N;
\end{equation}
or
\begin{equation}
\label{CSM}
\|C\|_1\to\min,\text{ subject to }Y=YC,\, \text{ diag}(C)=0;
\end{equation}
in the matrix form. It follows from the fundamental results from  \cite{CT}, \cite{D}, \cite{RV} that for matrices $Y$ with some
reasonable restrictions on $Y$ and for not very large Hamming weight of the ideal sparse solution, it can be uniquely found by solving convex problem (\ref{CS1}). There are other more efficient $\ell^1$-based methods for finding sparse solutions (e.g., see \cite{CWB},  \cite{PK2}, \cite{KP1}). 
\par
Provided that the data matrix is clean (maybe up to the noise in entries of not very high magnitude), the problem (\ref{CSM}) can be solved with the standard methods of Compressed Sensing. However, when some entries are corrupted or missed the special treatment of such input is necessary. We note that 
while.  when the the entries of the vector $\by_i$ are corrupted, problem (\ref{CS1}) can be efficiently solved with the algorithm from \cite{PK3}. However, that algorithm requires an uncorrupted matrix $Y$, what cannot be guaranteed in our case.
\par
The algorithm considered below does not require clean input data. We assume that those data may be corrupted with sparse errors, random noise is distributed over all vector entries, and a quite significant part of data is missing. 
\par
The property of errors to constitute a sparse set means that some (but not all) vector entries
are corrupted, i.e., those values are randomly replaced with different values or some random errors are added to the data entries. The locations (indexes) of the corrupted entries are unknown. 
\par
The second type of the corruption is missing data.
In information theory, the missed samples are called erasures. They have two main features.
First, the data values in erasures does not have any practical importance. The most natural way is to think about erasures as about lost data. Second, the coordinates of erasures are known. The second feature is the main difference with errors.
\par 
The noise is randomly introduced in each entry. Its magnitude is usually much less than the data magnitude. Two main sources of the noise are imperfect data measuring and finite precision of
data representation.
\par
It should be mentioned, that the matrices $Y$ in practical problems may be very far from the requirements  for the uniqueness of solutions. At the same time, the uniqueness of the solutions, as we mentioned above, is not necessary in our settings.  The error correction as well as erased entries recovery are also not crucial for the final goal of the clustering. We just wish to have the maximum of separation between indexes of 
the matrix $W$ corresponding to different subspaces. 
\par
In the case of successful clustering, the results for each $\cS^k$ may be used for further processing like
data noise removal, error correction, and so on. Such procedures become significantly more efficient when applied to low-rank submatrices of $Y$ corresponding to one subspace.
\par
Thus, in applications, the problems involving subspace clustering can be split into 3 stages: 
\par 
\begin{enumerate}
\item
 preprocessing (graph $W$ composing); 
\item search for clusters in the graphs;
\item  
processing on clusters.
\end{enumerate}
In this paper,  we develop a first stage algorithm helping to perform the second stage much more efficiently than the state-of-the-art algorithms. In \cite{PK5}, we designed the Greedy Sparse Subspace
Clustering (GSSC) algorithm relying on main principles of SSC algorithm from \cite{EV} but having increased clustering capabilities due to implementation of  greedy ideas at cost of the higher (about 5 times) computational complexity. Here we present an accelerated version of GSSC which is not slower
than SSC. As for its capability to separate subspaces, we will show it outperforms (sometimes significantly) not only SSC but even GSSC. We will call this algorithm the Fast GSSC or FGSSC.
\par
We do not discuss any aspects of improvements of stage 2.  We just 
take one of such algorithms, specifically the spectral clustering, and use it for comparison of the influence of our and competing preprocessing algorithms on the efficiency of clustering.
\par
As for stage 3, its content depends on an applied problem requesting subspace clustering.  The data recovery from incomplete and corrupted measurements is  one of typical possible goals of the third stage. Sometimes this problem  is called "Netflix Prize problem". We will briefly discuss below how the same problems of incompleteness and corruption can be solved within clustering preprocessing. However, for low-rank matrices (found clusters) it can be solved more efficiently. Among many existing algorithms we mention the most recent papers \cite{CR}, \cite{CJSC}
\cite{LCM}, \cite{WYGSM}, \cite{YYO}, \cite{PK4} providing the best results for input having both erasures and errors.
\par
It should be mentioned that in the case when $\dim(\sum \cS^k)<D$ the inverse order of steps can be applied.
On the step, the matrix $Y$ is completed and corrected with one of algorithms mentioned in the previous paragraph. Then the clean matrix $Y$ can be processed with SSC algorithm. Of coarse, this approach  fails when $d:=\dim(\sum \cS^k)$ is comparable or greater than $D$. It also is unreliable when we do not know the value $d$ in advance.

\par
In Section \ref{disc}, we discuss the  formal settings of optimization problem to be solved. In Sections \ref{SSC} we discuss the algorithm SSC (\cite{EV}) and our modification (\cite{PK5}) allowing to work with incomplete matrices.  In Section \ref{GSSC}, we describe the Greedy SSC (GSSC) algorithm constructed in \cite{PK5}  as external construction over SSC. In Section \ref{FGSSC}, we introduce the Fast GSSC  (FGSSC) algorithm which  is the main topic of this paper. The results of numerical experiments showing the consistency of the proposed approach  for both synthetic and real world data will be given in Section~\ref{exper}.
\par
 \section{Problem Settings}
 \label{disc}
We use main features of Sparse Subspace Clustering algorithm (SSC) from \cite{EV} 
for our  modification based on a greedy approach. Some additional extended reasoning related SSC can be found in  \cite{EV} and in the earlier paper \cite{EV0}.  The SSC algorithm in its last form created the tool for subspace  clustering resilient to data errors. In \cite{PK5}, we  adapted the SSC algorithm to the case when part of data is missing  allowing successful  clustering even if a significant fraction of data is unavailable. The same mechanism provides a~tool for more efficient error resilience.
\par
Very similar to SSC ideas of subspace self-representation  for subspace clustering were used also  in \cite{RTVM}. However, the error resilience mechanism in that paper works under assumption that there are enough uncorrupted data vectors.  This assumption is the case for the SSC algorithm selected as a foundation for GSSC.
\par
Optimization CS problems (\ref{CS1}) and  (\ref{CSM}) assume that the data are clean, i.e., they have no noise and errors. Considering the problem within the standard CS framework, in the presence of errors, the problem  \ref{CS} can be reformulated as finding the sparsest vectors $\bc$ (decomposition coefficients) and $\mathbf e$ (errors of "mesurements") satisfying the system of linear equations $\by=A\bc+\mathbf e$.
\par
It was mentioned in \cite{WYGSM} that the last system can be re-written as 
\begin{equation}
\label{errorcor}
\by=[A\, I]\left[\begin{array}{c} \bc \\ \mathbf e\end{array}\right],
\end{equation}
where $I$ is the identity matrix. Therefore, the problem of sparse reconstruction and error correction can be solved simultaneously with CS methods. In \cite{PK3}, we designed an algorithm efficiently finding sparse solutions to  (\ref{errorcor}).
\par
Unfortunately, the subspace clustering algorith cannot adopt this strategy straightforwardly because not only
"measurements" $\by_i$ are corrupted in (\ref{CS}) but "measuring matrix" $A=Y$ also can be corrupted.
It should be mentioned that if the error probability  is so  low that there exist uncorrupted columns of $Y$ constituting bases for all subspaces $\{\cS^k\}$, the method from \cite{PK3} can solve
the problem of sparse representation with simultaneous error correction. In what follows, the considered algorithm will admit a significantly higher error rate. In particular, all columns of $Y$ may be corrupted.
\par
Now we describe optimization settings accepted in this paper.
Following \cite{EV}, we introduce two (unknown for the~problem solver) $D\times N$  matrices $\hat E$ and $\hat Z$. The matrix $\hat E$ contains a sparse (i.e., $\#\{E_{ij}\ne 0\}<DN$) set of  errors with relatively large magnitudes.  The matrix $\hat Z$ defines the noise having a relatively low magnitude but distributed over all entries of $\hat Z$.
Thus, the clean data are representable as $Y-\hat E -\hat Z$. Therefore, when the data are corrupted with sparse errors and noise, the equation $Y=YC$ has to be replaces by 
\begin{equation}
\label{corr}
Y=YC +\hat E(I-C)+\hat Z(I-C).
\end{equation} 
The authors of \cite{EV} applied a reasonable simplification of the problem by replacing 2 last terms 
of (\ref{corr}) with some (unknown) sparse matrix $E:=\hat E(I-C)$ and the matrix with the deformed noise $Z=\hat Z(I-C)$. Provided that the sparse $C$ exists, the matrix $E$ still has to be sparse. This transformation  leads to some simplification of the optimization procedure. This is admissible simplification since, generally speaking, we do not need to correct and denoise the input data $Y$. Our only goal is  to find the sparse matrix $C$ which is a building block for the matrix $W$. Therefore, we do not need matrices $\hat E$ and $\hat Z$. 
\par
While there is an option to apply the error correction procedure  after subspace clustering, the problem of incorporation of this procedure, i.e., the matrices $\hat E$ and $\hat Z$, into subspace clustering still makes sense and deserves consideration in the future. However, it will be clear from what follows that  straightforward incorporation  leads to unjustified 
complexification of the optimization procedures.
\par
Taking into account the simplification from above reasoning, we can  formulate the constrained optimization problem 
\begin{equation}
\label{constr1}
\begin{gathered}
\min \|C\|_1+\lambda_e\|E\|_1+\frac{\lambda_z}{2}\|Z\|_{F}^2,
\\
\text{s.t. }Y=YC+E+Z, \,\text{diag}(C)=0.
\end{gathered}
\end{equation}
where $\|\cdot\|_F$ is the Frobenius matrix norm. If the clustering into affine subspaces is required, the additional constrain $C^T\mathbf 1=\mathbf 1$ is added.
\par
On the next step, using the representation $Z=Y-YC-E$ and introducing an auxiliary matrix 
$A\in\mathbb R^{N\times N}$, constrained optimization problem (\ref{constr1}) is transformed into
\begin{equation}
\label{constr2}
\begin{gathered}
\min \|C\|_1+\lambda_e\|E\|_1+\frac{\lambda_z}{2}\|Y-YA-E\|_{F}^2
\\
\text{s.t. } A^T\mathbf 1=\mathbf 1,\, A=C-\text{diag}(C).
\end{gathered}
\end{equation}
Optimization problems  (\ref{constr1})  and (\ref{constr2})  are equivalent.
Indeed, obviously, at the point of the extremum of (\ref{constr2}), diag$(C)=0$. Hence, $A=C$.
\par
At last, the quadratic penalty functions with the~weight $\rho/2$ corresponding to constrains are added to
the functional in  (\ref{constr2}) and the Lagrangian functional is composed. The final Lagrangian functional is as follows
\begin{equation}
\label{lagr}
\begin{gathered}
\mathcal L(C,A,E, \boldsymbol \delta, \Delta)=\min \|C\|_1\\+\lambda_e\|E\|_1 +\frac{\lambda_z}{2}\|Y-YA-E\|_{F}^2
\\
+\frac{\rho}{2}\|A^T\mathbf 1-\mathbf 1\|_2^2+\frac{\rho}{2}\|A-C+\text{diag}(C))\|_F^2
\\
+\boldsymbol\delta^T(A^T\mathbf 1-\mathbf 1)+\tr(\Delta^T(A-C+\text{diag}(C))),
\end{gathered}
\end{equation}
where the vector $\boldsymbol \delta$ and the matrix $\Delta$ are Lagrangian multipliers.
Obviously, since the penalty functions are formed from the constrains, they do not change the point and the value of the minimum. 
\par
The first terms in lines 3 and 4 of (\ref{lagr}) have to be removed when only linear subspace  clusters are considered.
\section{Algorithms}
\label{alg}
\subsection{Sparse Subspace Clustering Algorithm}
\label{SSC}
For finding the stationary point of functional (\ref{lagr}) an Alternating Direction Method  of Multipliers (ADMM,  \cite{BPCPE}) is used. In \cite{EV}, this procedure is a crucial part of the entire algorithm which is called the Sparse Subspace Clustering algorithm.  While this minimization constitutes only a part of the entire SSC algorithm, for the sake of brevity, we will call it the  SSC algorithm.  Before formal description of the SSC algorithm we discuss the selection of parameters in (\ref{lagr}).
\par
The parameters $\lambda_e$ and $\lambda_z$ in (\ref{lagr}) are selected in advance. They define the compromise
between good approximation of $Y$ with $YC$ and the high sparsity of $C$. The general rule is 
to set the larger values of the parameters for the less level of the noise or errors. In \cite{EV}, the selection of the parameters 
by formulas
\begin{equation}
\label{pars}
	\lambda_e=\alpha_e/\mu_e,\qquad \lambda_z=\alpha_z/\mu_z,
\end{equation}
where $\alpha_e,\, \alpha_z>1$ and 
$$
\mu_e:=\min_i\max_{j\ne i}\|\by_j\|_1,\quad\mu_z:=\min_i\max_{j\ne i}|\by_i^T\by_j|,
$$
is recommended.
\par
The initial parameter $\rho=\rho^0$ is set in advance. It is updated as $\rho:=\rho^{k+1}=\rho^k\mu$ with iterations of SSC algorithm. We notice that, adding the penalty terms, we do not change the problem. It still has the same minimum. However,
the appropriate selection of $\mu$ and $\rho^0$ accelerates the algorithm convergence significantly.
\par
We will need the following notation
$$
 S_{\epsilon}[x]:=\left\{\begin{array}{ll} x-\epsilon, &x>\epsilon,\\x+\epsilon, &x<-\epsilon,\\0, &\text{otherwise};\end{array}\right.
$$
where $x$ can be either a number or a vector or a matrix. The operator $S_{\epsilon}[\cdot]$ is called the shrinkage operator. 
\par
In what follows, we accept that the data (matrix $Y$) is available only at entries with indexes on the set 
$\Omega\subset \{1,\dots,D\}\times\{1,\dots,N\}$. $\chi_\Omega$ is a characteristic function of the set $\Omega$. The symbol $\odot$ will be used for the entrywise products of matrices.
\par
\begin{algorithm}[h]
\label{asdfg}
\caption{{\bf SSC} (modified)}\label{SSCalg} 
{\bf Input: } $Y\in\mathbb R^{D\times N}$, $\Omega\in\mathbb R^{D\times N}$.
\begin{algorithmic}[1]
\State {\bf Initialization:} $ C^0:=0$, $A^0:=0$, $E^0:=0$, $\boldsymbol\delta^{0}:=0$, $\Delta^0:=0$, $k:=0$,   STOP:=false, $\epsilon>0$, $\rho^0>0$, $\mu\ge1$
\While{STOP==false}
\State update $A^{k+1}$ by solving the system of linear equations
\begin{multline}
\label{a}
(\lambda_zY^TY+\rho^kI+\rho^k\mathbf 1\mathbf 1^T)A^{k+1}\\=\lambda_zY^T(Y-E^k)
+\rho^k(\mathbf 1\mathbf 1^T+C^k)-\mathbf 1\boldsymbol\delta^{kT}-\Delta^k.
\end{multline}
\State update 
$$
C^{k+1}:=J^k-\diag(J^k),
$$
where
$
J^k:=S_{{1}/{\rho^k}}[A^{k+1}+\Delta^k/\rho^k].
$
\State update
\begin{multline}
\label{e}
E^{k+1}:=\chi_\Omega \odot S_{\frac{\lambda_e}{\lambda_z}}[Y-YA^{k+1}]+\\
(1-\chi_{\Omega })\odot (Y-YA^{k+1})
\end{multline}
\State  update
$$
\boldsymbol\delta^{k+1}:=\boldsymbol\delta^{k}+\rho^k(A^{k+1}\mathbf 1-\mathbf 1),
$$
\State update
$$
\Delta^{k+1}:=
\Delta^{k}+\rho^k(A^{k+1}-C^{k+1}).
$$
 \If{
$$
\|A^{k+1}\mathbf 1-\mathbf 1\|_\infty<\epsilon,\quad\&\quad
\|A^{k+1}-C^{k+1}\|_\infty<\epsilon, 
$$
\&
$$
\|A^{k+1}-A^k\|_\infty<\epsilon,\quad\&\quad
\|E^{k+1}-E^k\|_\infty<\epsilon
$$
}
STOP=true;
\EndIf
\State update {$\rho^{k+1}:=\rho^k\mu$}

\State update $k:= k+1$

\EndWhile

\Return $C^*=C^{k}$, $Y_{out}:=Y-E^k$.
\end{algorithmic}
\end{algorithm}
\par
  Each iteration of the  algorithm is based on consecutive optimization with respect to each of the unknown values $A$, $C$, $E$, $\boldsymbol\delta$, $\Delta$ which are initialized by zeros before the algorithm starts. 
\par
Due to an appropriate form of functional (\ref{lagr}), optimization  of $A^k$, $C^k$, and $E^k$ in Algorithm  \ref{SSCalg} is simple and computationally efficient. Moreover, lines 3--5 give the optimal (for the fixed other variables) solution in the explicit form. The five formulas (lines 3--7) for updating the unknown values are discussed below.
\par
The matrix $A^{k+1}$ is a solution of the matrix equation with the unknown matrix of size $N\times N$. However, when $D<N$, the complexity of this operation is below $O(N^3)$. Indeed,  obviously,

\begin{equation}
\label{aa}
(\lambda_zY^TY+\rho I+\rho \mathbf 1\mathbf 1^T)^{-1}={\rho^{-1}}(I+\tilde{Y}^T Q^{-1}\tilde{Y}),
\end{equation}
where 
$$
\tilde Y=\left\{\begin{array}{ll} Y, & \text{case of linear subspaces;}\\ 
 \left[\begin{array}{c} Y \\ \mathbf 1^T\end{array}\right], & \text{case of affine subspaces;}\end{array}\right.
$$
$$
Q= V+\tilde Y\tilde Y^T\in\mathbb R^{D\times D}\,(\text{or } \mathbb R^{(D+1)\times (D+1)}),
$$
$V$ is a diagonal matrix, $V_{ii}=\rho/\lambda_z$, $i=1,\dots,D$; $V_{D+1,D+1}=1$ for the case of affine subspaces. 
\par
In typical cases, the number of points $N$ is much greater than the dimension  of the ambient
space $D$. Thus, the complexity of the matrix inversion  in (\ref{aa}) is $O(D^3)$. Computation of the matrix $Q$ requires $O(D^2N)$ operations but, in fact, computing $YY^T$ has to be performed only on the first iteration of Algorithm~\ref{SSCalg}.  Unfortunately, the follow-up matrix multiplication has complexity $O(DN^2)$. Nevertheless, provided that  $D\ll N$, computing of the solution to  (\ref{a}) may be much faster than the straightforward solving that system. 
\par
Lines 4 and 5 of Algorithm \ref{SSCalg} represents explicit optimization of the functional (\ref{lagr}) with respect to $C$ and $E$ correspondingly, provided that other variables are fixed.
\par
Lines 6 and 7 present updates of the Lagrangian multipliers $\boldsymbol\delta$ and $\Delta$. Those updates do not solve any optimization problems. Their intention is to move the value of the Lagrangian
functional toward it minimum value. More derailed arguments and discussion related to Alternating Direction Method of Multipliers justifying this step can be found in \cite{BPCPE}.
\par
Algorithm \ref{SSCalg} is a modified version of the original SSC algorithm from \cite{EV} which
 gave the state-of-the-art benchmarks for subspace clustering 
problems. 
\par
Our first modification consists in taking into account missing data by means of replacing
the update formula $E^{k+1}:= S_{\frac{\lambda_e}{\lambda_z}}[Y-YA^{k+1}]$ with our version (\ref{e}). 
\par
To make clear how that modification use the a priori knowledge of the  coordinates of erased entries,
let us consider the mechanism of the influence of the value $\lambda_e$ on the  output matrix $C$.  The parameter $\lambda_e$ sets
the balance between the higher level of the sparsity of $C$ with the more populated error matrix
$E$ vs. the less sparse $C$ and the less populated matrix $E$. 
Setting too small $\lambda_e$  allows too many "errors" and very sparse $C$. However,  probably, this is not what we want. This would mean that sake of $C$ sparsity we introduced too large distortion into the input data $Y$. At the same time, if we know for sure or almost for sure that some 
entry of $Y$ with coordinates $(i, j)$  is corrupted, we  loose nothing by assigning to this element an individual small weight in functional (\ref{lagr}). This weight can be much less than $\lambda_e$ or even equal to $0$. Thus, we have to replace the term $\|E\|_1$ with $\|\chi_\Omega\odot E\|_1 $. This means that in formula (\ref{e}) we  apply different shrinkage threshold for different indexes.
Generally speaking, it makes sense to use all range of non-negative real numbers to reflect our knowledge about $E$. Say, highly reliable entries have to be protected from distortion by the weight greater than 1. However, in this paper we restrict ourself with two-level entries:
either 1 (no knowlege) or $0$ (erasure or entries suspicious to be  errors), using the characteristic function $\chi_\Omega$ for reweighting.  
\par
In the case when no erasures are reported, we have  $\chi_\Omega\equiv1$. Therefore, Algorithm~1 at line 5 works like a regular SSC algorithm. 
\par
Our second modification of SSC consists in updating $\rho$ (line 10). As we can judge, no update was used in the original algorithm in \cite{EV}. While such update does not change the optimal value of the functional, it brings some algorithm acceleration.
\par
In Section \ref{exper}, we will give numerical evidences of the efficiency of Algorithm \ref{SSCalg} for
fighting the problem of missing samples. 
\subsection{Greedy Sparse Subspace Clustering}
\label{GSSC}
\par
While  the original SSC algorithm in \cite{EV} did not have any special tool against missing samples, it engaged the error resilience mechanism. In particular, it can perform clustering on data with some restricted number of erasures. However, by our opinion, the potential power of the error resilience laid  in Algorithm \ref{SSCalg} was no not realized completely. 
\par
The main idea lying in the foundation of our follow-up reasoning is a simple   information theory principle. Assume that in the beginning no information about errors in data is available. Then if we are able (say, using one run of the SSC algorithm) to identify that some data entries contain errors, the errors can be re-qualified into erasures (marked as erasures) and run the algorithm working efficiently with erasures again.  Our intuition bases on information theory principles tells us that any extra knowledge (say, coordinates of errors) has to give some benefits  for algorithm.  At the same time, this strategy has some restrictions. Let us discuss 
how this trick may change (hopefully improve) our algorithm. First of all, we have to be sure that 
we found and marked actual errors. Moving "healthy" entries into the list of erasures, we destroy correct information and reduce the algorithm ability to make reliable conclusions. Thus, the procedure of finding the error locations has to be reliable. 
However, even if we are able to find correct error locations, we have to take into account that re-qualification of errors
into erasures implies the necessity to accept that the value at the erroneous entries do not contain 
any useful information. Such claim is true only if the values at error locations are independent from each other and from the correct values. The most typical case when those independence requirements fail is the mixture of data with noise. While all entries are corrupted, they still have information about
data which can be used for data recovery. At the same time, moving a data entry into the list of erasures we lose that useful information. Thus, the idea of moving the errors into the list of erasures 
may bring benefits only when the algorithm of finding  error locations is consistent and amount 
of useful information in corrupted entries is not too large. 
\par
 In spite of the mentioned above restrictions, in many cases,
incomplete information can be processed more efficiently than erroneous information.  
\par
Our suggestion is to attract ideas of greedy algorithms to increase the capability of the SSC algorithm in
subspace clustering.  Greedy algorithms are very popular in non-linear approximation (especially in redundant systems) when the global optimization is replaced with iterative selection of the most probable candidates
from the point of view of their prospective contribution into approximation. The procedure is repeated 
with selection of new entries, considering the previously selected entries as reliable  with guaranteed participation in approximation. The most typical case is Orthogonal Greedy Algorithm (OGA) consisting in selection of the approximating entries having the biggest inner products with the current approximation residual and follow-up orthogonal projection of the approximated object onto the span 
of the selected entries. 
\par
In many cases, OGA allows to find the sparsest representations if they exist. In \cite{KP1} and \cite{PK2}, we applied the greedy idea in combination with the reweighted $\ell^1$-minimization to CS problem of finding the sparsest solutions of underdetermined system.   We used the existing $\ell^1$-minimization scheme from \cite{CWB} with the the opportunity to reweight entries. When the basic algorithm fails, the greedy block picks locations of the biggest (the most reliable) entries in the decomposition whose magnitudes
are higher than some threshold. Those entries are considered as reliable. Therefore, they get the less
weight in the $\ell^1$-norm while other entries are competing on next iterations for the right to be picked up. 
\par 
The similar idea was employed in our recent paper \cite{PK4}, where the greedy approach was applied 
to the algorithm  for completion of low-rank matrices from incomplete highly corrupted samples from \cite{LCM} based  on  the Augmented Lagrange Multipliers method. The simple greedy modification of
the matrix completion algorithm from \cite{LCM} gave the boost in the algorithm restoration capability.
\par
Now we discuss details how the greedy approach can be incorporated in (to be more precise {\it over}) the SSC algorithm (cf. \cite{PK5}).  As above, we use the set $\Omega$ for keeping the information about erasures. However, we also will use it as a storage of  information about coordinates of presumptive errors. Such information will be extracted from the error matrix $E$. Thus, $\chi_\Omega$ vanishes at the selected in advance erasures and at the points suspicious to be errors.
\par
The entries which are suspicious to be  errors  are dynamically removed from $\Omega$ after each iteration of the greedy algorithm.
\par
\begin{algorithm}[h]
\caption{{\bf GSSC}}\label{GSSCalg} 
{\bf Input: } $Y\in\mathbb R^{D\times N}$, $\Omega\in\mathbb R^{D\times N}$.
\begin{algorithmic}[1]
\State {\bf Initialization:} $ Y^0:=Y\odot\chi_\Omega$, $\Omega^0=\Omega$,  $ 0<\alpha_1,\alpha_2,\beta<1$, maxIter$>0$, $k:=0$,   STOP:=false.
\State $\text{mxMed}:=\max_{1\le j \le N}\text{\,median}\,(|\by_j|)$;
\While{STOP==false}
\State $ (C^k, Y_{out})=\text{SSC}(Y^k, \Omega^k )$;
\State $E= |Y_{out}-Y^k|$;
\If {$k==0$}
\State $M:=\max\{\alpha_1 \cdot\max\{E\}, \alpha_2\cdot\text{mxMed}\}$;     
\EndIf
\State
$M:=\beta M$;
\State
$\Omega^{k+1}:=\Omega^k\setminus \{(i,j)\mid E_{i,j}>M\}$;
\State
$Y^{k+1}:=\chi_{\Omega^{k+1}}\odot Y^k+(1-\chi_{\Omega^{k+1}})\odot Y_{out}$;
\If{$k==\text{maxIter}$}
\State STOP:=true;
\EndIf
\State $k:=k+1$;
\EndWhile
\end{algorithmic}
\par\noindent{\bf Output:} $C^{k-1}$.
\end{algorithm}
Thus, in the Greedy Sparse Subspace Clustering (GSSC), we organize an external loop over the SSC.
\par
One iteration of our greedy algorithm consists in running the modified version of SSC and  
$\Omega$ and $Y$ updates.

While $E^n$ in Algorithm \ref{SSCalg} is not a genuine matrix of errors,  this is not serious drawback for the original SSC algorithm. However,
for GSSC this may lead to unjustified and very undesirable shrinkage of the set $\Omega$. A more accurate estimate of the error set in future algorithms may bring significant benefits for GSSC. 
 
\par
\subsection{Fast Greedy Sparse Subspace Clustering Algorithm}
\label{FGSSC}
Now we present the Fast Greedy Sparse Subspace Clustering (FGSSC) algorithm (see Algorithm 3) which is main contribution of this paper.
The FGSSC consists in incorporation of the greedy update of the set $\Omega$ into Algorithm \ref{SSCalg}. 
\par
Except for the obvious elimination of the external GSSC loop leading to the acceleration of the algorithm, accurate tuning of the FGSSC parameters
brings quite significant increase of the algorithm capability. 

\begin{algorithm}[h]
\caption{{\bf FGSSC}}\label{FGSSCalg} 
{\bf Input: } $Y\in\mathbb R^{D\times N}$, $\Omega\in\mathbb R^{D\times N}$.
\begin{algorithmic}[1]
\State {\bf Initialization:} $ C^0:=0$, $A^0:=0$, $E^0:=0$, $\boldsymbol\delta^{0}:=0$, $\Delta^0:=0$, $k:=0$,  $M=\infty$, STOP:=false, $\epsilon>0$, $\rho^0>0$, $\mu\ge1$
\State mxMed=median$(\chi_\Omega\odot Y)$;
\While{STOP==false} 
\If{ $k==k_0$ }
       $M:= \alpha_0\,\|E_{ij}\|_\infty$;
 \Else
       \If{k is even}
       \State  $M:=\max\{\alpha_1M,\alpha_2\text{\,mxMed}\}$;
       \EndIf
  \EndIf
   
\State   $ \Omega:=\Omega \setminus \{(i,j)\mid |E_{i,j}|>M\}$;
\State update $A^{k+1}$ by solving the system of linear equations
\begin{multline*}
(\lambda_zY^TY+\rho^kI+\rho^k\mathbf 1\mathbf 1^T)A^{k+1}\\=\lambda_zY^T(Y-E^k)
+\rho^k(\mathbf 1\mathbf 1^T+C^k)-\mathbf 1\boldsymbol\delta^{kT}-\Delta^k.
\end{multline*}

\State update $C^{k+1}:=J^k-\diag(J^k)$,
where
$
J^k:=S_{{1}/{\rho^k}}[A^{k+1}+\Delta^k/\rho^k].
$
\State update
\begin{multline*}
E^{k+1}:=\chi_\Omega \odot S_{\frac{\lambda_e}{\lambda_z}}[Y-YA^{k+1}]+\\
(1-\chi_{\Omega })\odot (Y-YA^{k+1})
\end{multline*}
\State  update
$
\boldsymbol\delta^{k+1}:=\boldsymbol\delta^{k}+\rho^k(A^{k+1}\mathbf 1-\mathbf 1),
$
\State update
$
\Delta^{k+1}:=
\Delta^{k}+\rho^k(A^{k+1}-C^{k+1}).
$
 \If{
$
\|A^{k+1}\mathbf 1-\mathbf 1\|_\infty<\epsilon,$ \&\,
$\|A^{k+1}-C^{k+1}\|_\infty<\epsilon, 
$
\& \,
$
\|A^{k+1}-A^k\|_\infty<\epsilon,\,$\&\,
$\|E^{k+1}-E^k\|_\infty<\epsilon
$
}
STOP:=true;
\EndIf

\If {$k$ is even}
\State         $Y:=Y- \chi_\Omega E^k$;
 \State         $E^{k+1}:=0$;
\State update $\mu_e(\Omega)$ and $\mu_z(\Omega)$;
\EndIf
\State update {$\rho^{k+1}:=\rho^k\mu$}
\State update $k:= k+1$
\EndWhile
\par\noindent
\Return $C^*:=C^{k}$
\end{algorithmic}
\end{algorithm}
\par
Now we give comments on Algorithm \ref{FGSSCalg} implementation. 
\par
FGSSC starts with a few  iteration with blocked update of the set $\Omega$.  A very large value is set at the threshold for update
$M$. It  gets a realistic value (see line 4) after  $k_0$ iterations. Those initial $k_0$-step tuning allows  to fill erasures with some reasonable values approximating "genuine" values of $Y$. Those steps also
provide us with an estimate of the largest values of errors. 
\par
The algorithm iterations are grouped in pairs. After even iterations the estimate of the error matrix $E$ accumulated for 2 iteration is subtracted  from the data matrix $Y$ (line 18). After  that, $E$ is set to 0. 
\par
Updates of  $\mu_e(\Omega)$ and $\mu_z(\Omega)$ in line 21 are optional they makes a sense when $\Omega$ has significant change during iterations and erroneous entries may have large magnitudes.
\par
The number of erasures (and even their coordinates) is known before the processing. So some of the algorithm parameters can be tuned, according to the known information.
The parameters $\alpha_e$, $\alpha_z$ can be used for the algorithm  fine adjustment to the erasure density of the input data.

\section{Numerical Experiments} 
\label{exper}

In the beginning, we will present the comparison of the FGSSC and SSC algorithms on two types of synthetic data. The third part of this section is devoted to the problem of the face recognition. To be more precise we consider face images classification problem and present comparison of the FGSSC and SSC algorithms. 
\par
For the paper size reduction, we do  not present  the results of GSSC algorithm which can be found in our paper \cite{PK5}. We just mention that it provides the clustering efficiency approximately in the middle between SSC and FGSSC and its execution time is 2-3 times greater than that time for FGSSC.
\par
We also have to emphasize that in this paper we do not try to intrude into the spectral clustering algorithm. We just provide equal opportunity for the SSC and FGSSC algorithms for the final cluster selection based on the matrices $C$ obtained by each of algorithms. For this reason we do not discuss
here some important topics like {\it data outliers}.
\subsection{Synthetic Input I}
\label{synth1}
\par
The input data for the first experiment was composed in accordance with the model given in \cite{EV}. 105 data vectors of dimension $D=50$ are equally split between  three 
4-dimensional linear spaces $\{\cS^i\}_{i=1}^3$. To make the problem more complicated each of those 3 spaces belongs to sum of two others. The smallest angles between spaces $\cS^i$ and $\cS^j$ 
are defined by formulas 
$$
\cos \theta_{ij}=\max_{\bu\in\cS^i,\,\bv\in\cS^j} \frac{\bu^T\bv}{\|\bu\|_2 \|\bv\|_2},\,i,j=1,2,3.
$$
We construct the data sets using vectors generated by decompositions with random coefficients in orthonormal bases $\be^j_i$ of the spaces $\cS^j$. Three vectors $\be^j_1$ belong to the same  2D-plane with angles $\widehat{\be^1_1 \be_ 1^2}=\widehat{\be^2_1 \be_ 1^3}=\theta$ and 
$\widehat{\be^1_1 \be_ 1^3}=2\theta$. The vectors $\be_2^1, \be_2^2, \be_3^1, \be_3^2, \be_4^1, \be_4^2$ are 
mutually orthogonal and orthogonal to $\be_1^1, \be_1^2, \be_1^3$; 
$\be_j^3=(\be_j^1+\be_j^2)/\sqrt{2}$, $j=2,3,4$.
The generator of standard normal distribution is used to generate data decomposition coefficients.
After the  generation, a random unitary matrix is applied to the result to avoid zeros in some regions of the matrix $Y$. 
\par
We use the notation $P_{ers}$  and $P_{err}$ for probabilities of erasures and  errors correspondingly.
\par
When we generate erasures we set random entries of the matrix $Y$ with probability $P_{ers}$ to zero since no a priori information about those values is known.
\par
The coordinates of samples with errors are generated randomly with probability $P_{err}$.
We use the additive model of errors,  adding  values of errors to the correct entries of $Y$.
The magnitudes of errors are taken from  standard normal distribution. Obviously, as we discussed in Section \ref{GSSC}, for such additive model the~corrupted entries  contain a lot of useful information about the data and potentially may be processed  better than with our FGSSC algorithm. However, in spite of that,  we will see below that improvement of  clustering efficiency  over SSC is significant.
\par
We run 50 trials of FGSSC and SSC algorithms  for each combination of $(\theta, P_{err}, P_{ers})$, $$0\le\theta\le 60^\circ,$$ 
$$0\le P_{err}\le0.5,$$  
$$0\le P_{ers}\le0.7,$$ 
and output average values of misclassification. We note that for the angle $\theta=0$ the spaces $\{\cS^l\}$ have a common line and $\dim (\oplus_{l=1}^3\cS^l)=7$. Nevertheless, we will see that SSC and especially GSSC shows high capability  even for these hard settings.
\par
\noindent
\begin{picture}(00,110)
\put(0,5){          \epsfig{file=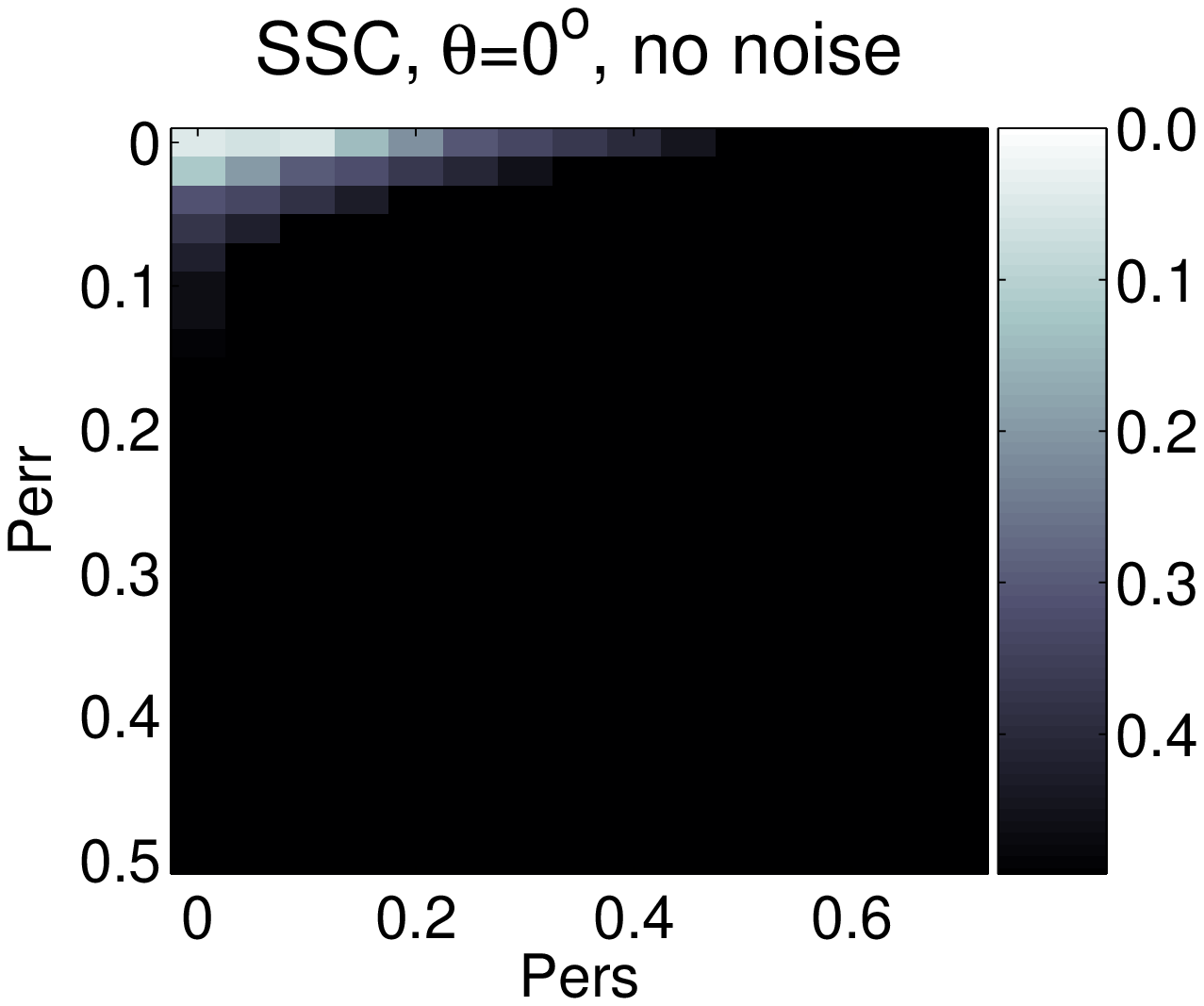,
           height=100pt, width=130pt}
		  } 
\put(120,5){          \epsfig{file=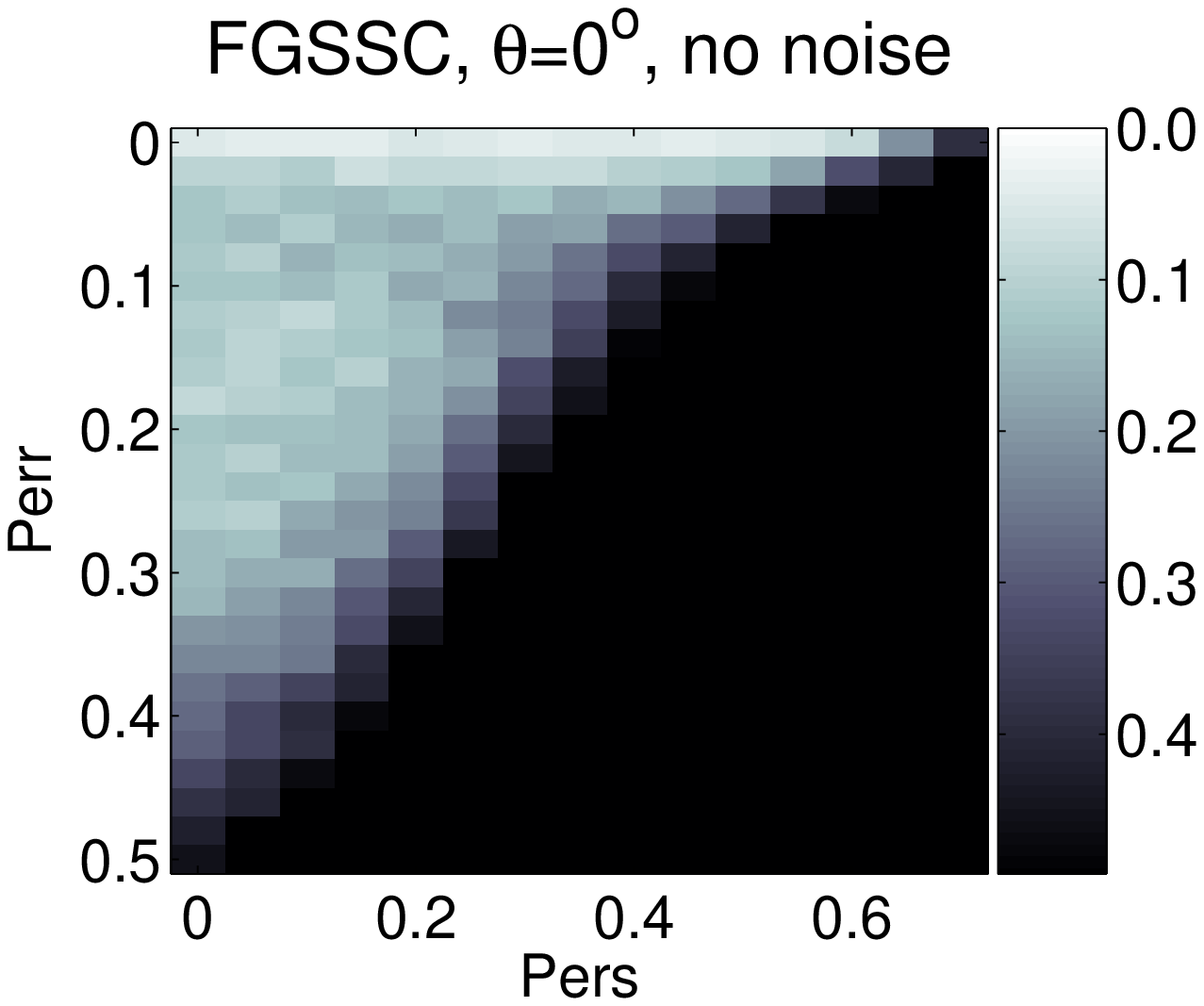,
           height=100pt, width=130pt}
		  } 
\end{picture}
\par
\noindent
\begin{picture}(00,110)
\put(0,5){          \epsfig{file=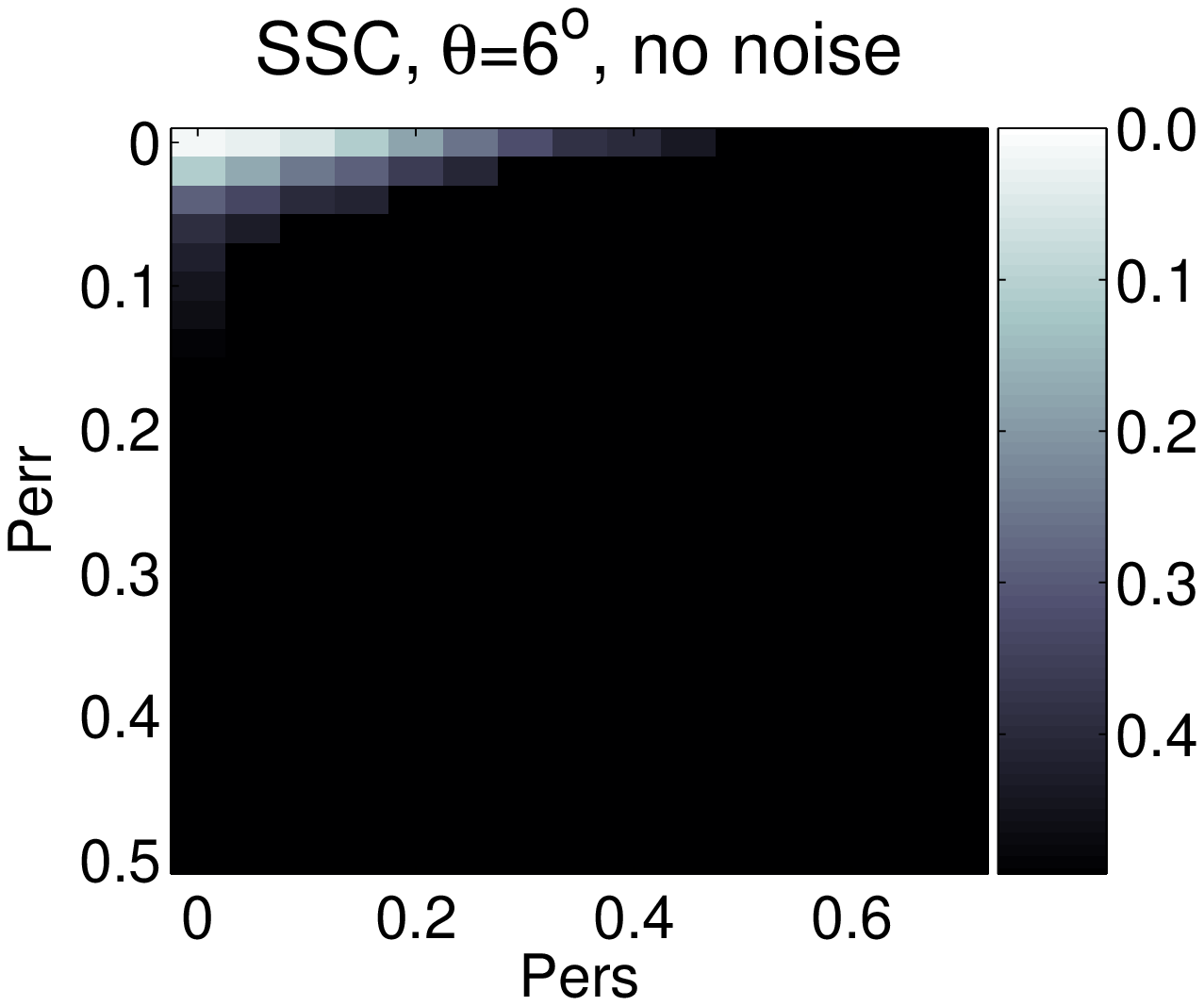,
           height=100pt, width=130pt}
		  } 
\put(120,5){          \epsfig{file=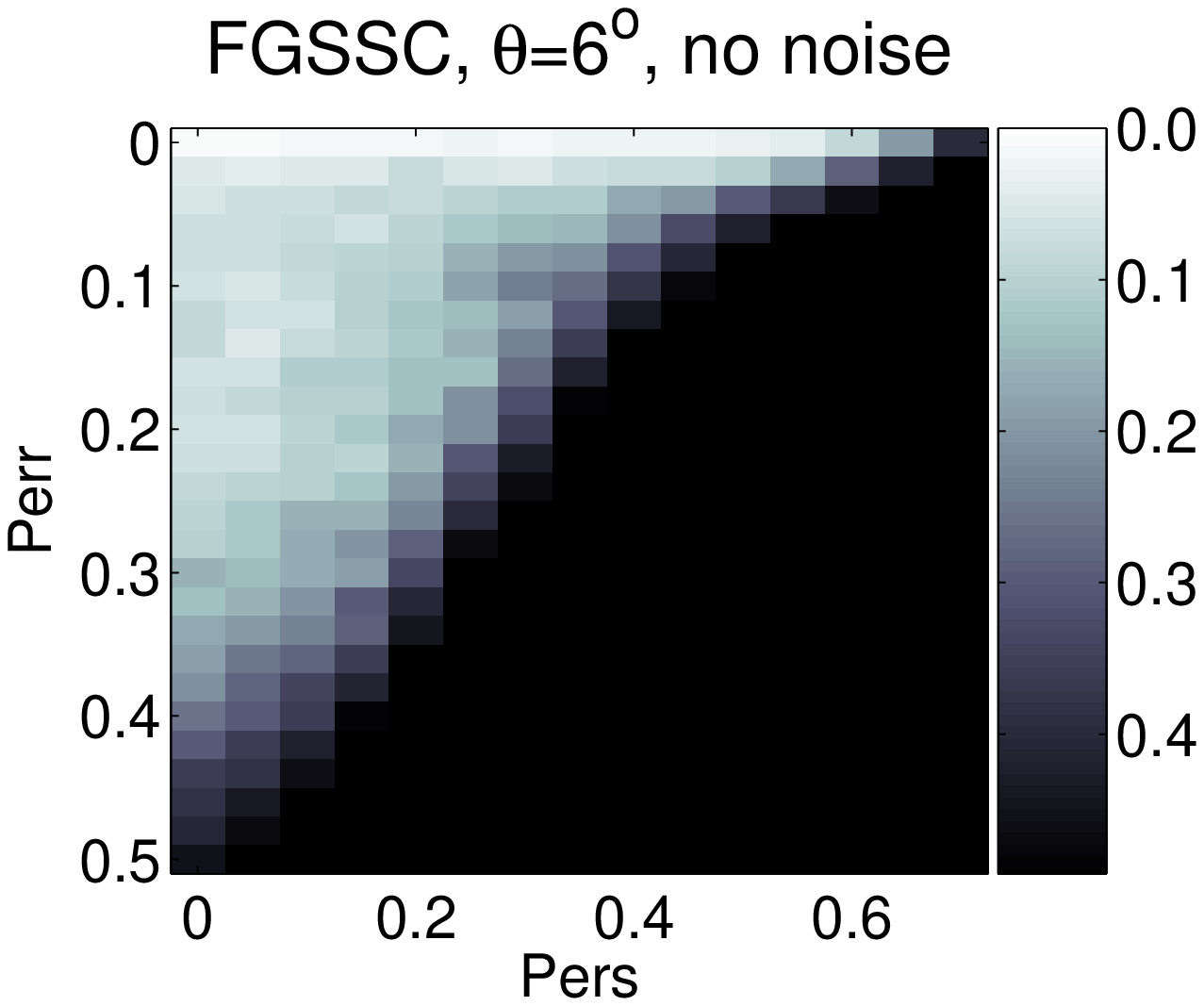,
           height=100pt, width=130pt}
		  } 
\end{picture}
\par
\noindent
\begin{picture}(00,115)
\put(0,10){          \epsfig{file=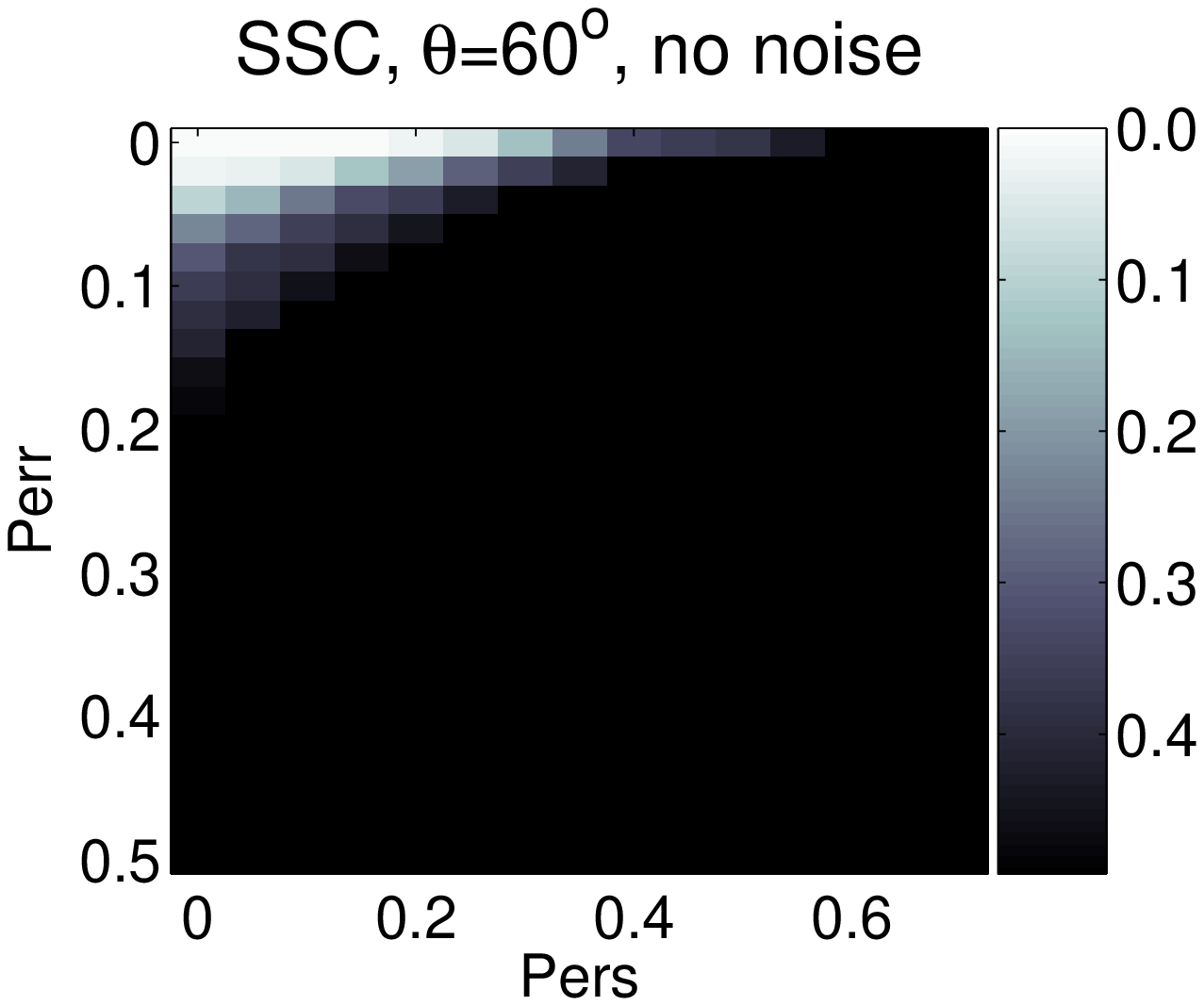,
           height=100pt, width=130pt}
		  } 
\put(120,10){          \epsfig{file=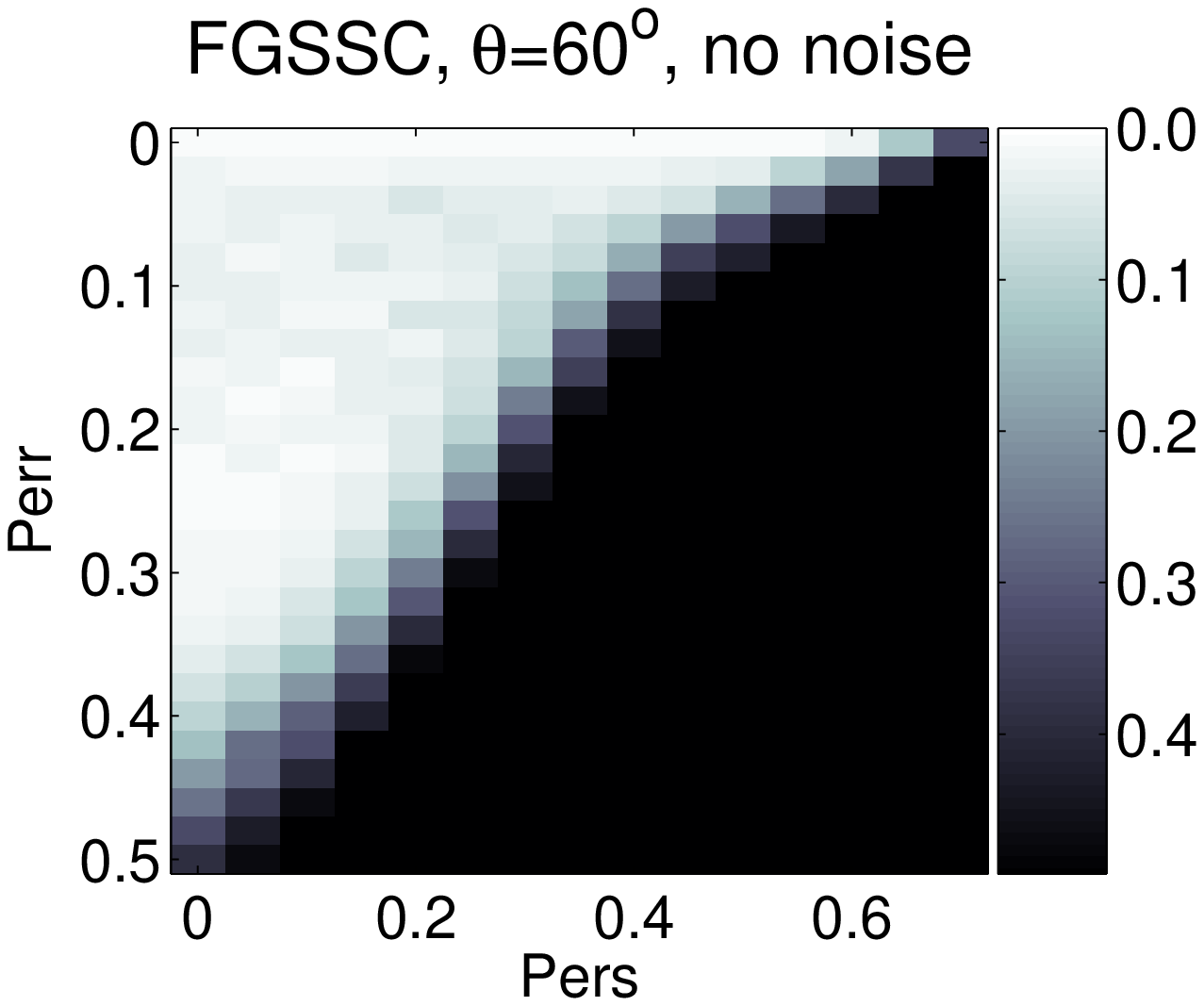,
           height=100pt, width=130pt}
		  } 
\put(70, 0){\footnotesize{\bf Fig.1. Clustering on noise free input.} }
\end{picture}
\par\noindent

\par
Now we describe the algorithm parameters. 
\par
SSC processing was performed with $\alpha_e$ linearly changing from 5 (no erasures) to 24 for 0.7 erasures. $\alpha_z=7$, $\rho^0=10$, $\mu=1.05$, $\epsilon=0.001$. 
\par
The results for the FGSSC algorithm presented on Fig. 1 were obtained with parameters $\alpha_0=0.6$, $\alpha_1=0.95$, $\alpha_2=1.0$; the parameter $\alpha_e$ is linearly changed from $11$ for $P_{ers}= 0.0$ to 22 for $P_{ers}= 0.7$; $\alpha_z=20$. 
\par
 Each point of the images is obtained as average value over 50 trials when both the matrix Y, sets of erasures and errors as well as the values of errors are randomly drawn, according to the models described above. The intensity of each point takes a range from "white", when there is no errors in clustering, to "black", when
more than 50\% of vectors were misclassified.
\par
The results confirms that FGSSC has much higher error/erasure resilience than SSC. For all models of input data and for both algorithms "the phase transition  curve"  is observed.
For the case $\theta=0^\circ$, clustering cannot be absolutely perfect even for FGSSC.  Indeed, the clustering for $P_{err}>0$ and $P_{ers}>0$ cannot be better than for error free model. At the same time, FGSSC was designed for better error handling. In the error free case, FGSSC has no advantage over the SSC algorithm.  Thus, the images on Fig.~1 for $\theta=0$ have the gray background of the approximate level $0.042$ equal to the rate of misclassification of SSC.  We believe that the reason of the misclassification lies in the method how we define the success. For $\theta=0^\circ$, 
there is a~common line (a 1D subspace) belonging to all subspaces $\cS^l$. For  points close to that line,  the considered algorithm has to make a hard decision, appointing only one cluster for each such point. 
Probably, for most of applied problems, the information about multiple subspaces accommodation is more useful than the unique space selection.
We advocate for such multiple selection because the typical follow-up problem after clustering is 
correction of errors in each of clusters. For this problem, it is not important to which of clusters the vector belonged from the beginning. When the vector affiliation is really important, side information has to be attracted.
\par
The second part of experiment deals with noisy data processing.  Independent Gaussian noise
of magnitude 10\% of mean square value of the data matrix $Y$ (i.e., the noise level is -20dB) is  applied to the matrix $Y$. On Fig. 2, we present the results of processing of the noisy input analogous to results on Fig. 1. Evidently, that this quite strong noise has minor influence on the clustering efficiency. 
\par
If we increase the  noise up to -15 dB, the algorithms still resist. For -10 dB (see Fig. 3) FGSSC  looses a  lot but still significantly outperforms SSC.  Those losses are obviously  caused by the increase of the noise fraction 
in the~mixture errors-erasures-noise, while the greedy idea efficiently works for highly localized corruption like 
errors and erasures.

\par
\noindent
\begin{picture}(00,110)
\put(0,5){          \epsfig{file=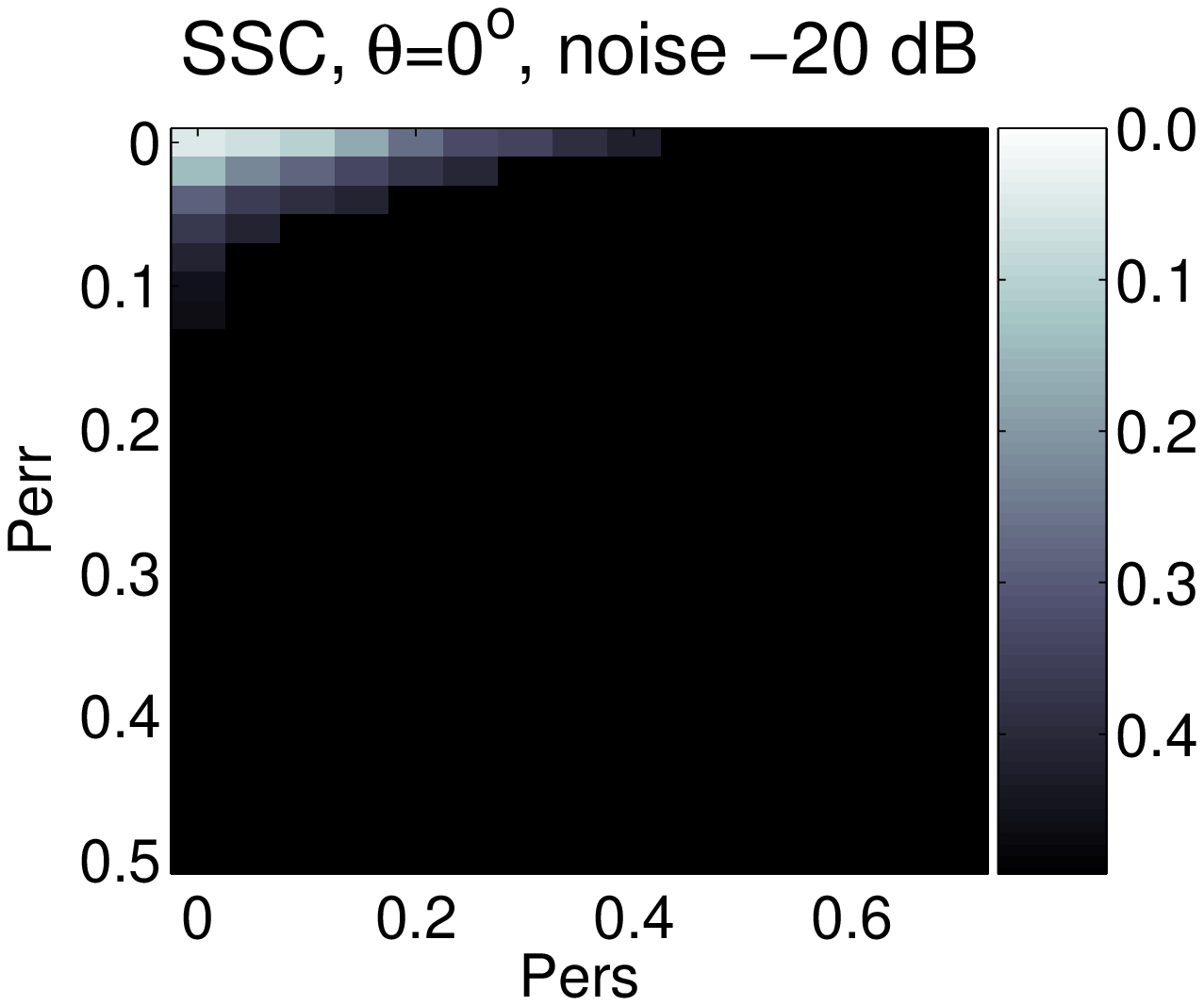,
           height=100pt, width=130pt}
		  } 
\put(120,5){          \epsfig{file=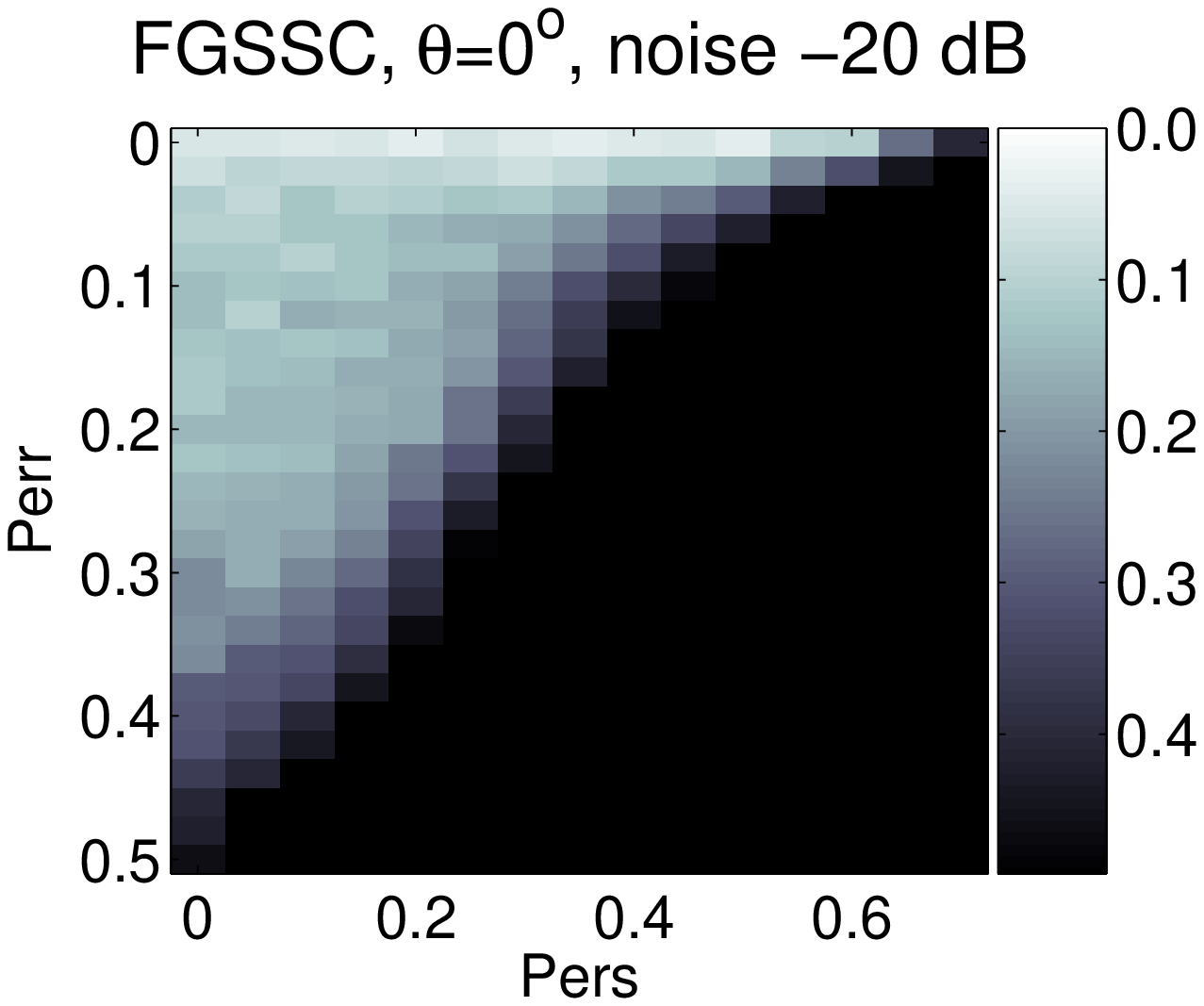,
           height=100pt, width=130pt}
		  } 

\end{picture}
\par
\noindent
\begin{picture}(00,110)
\put(0,5){          \epsfig{file=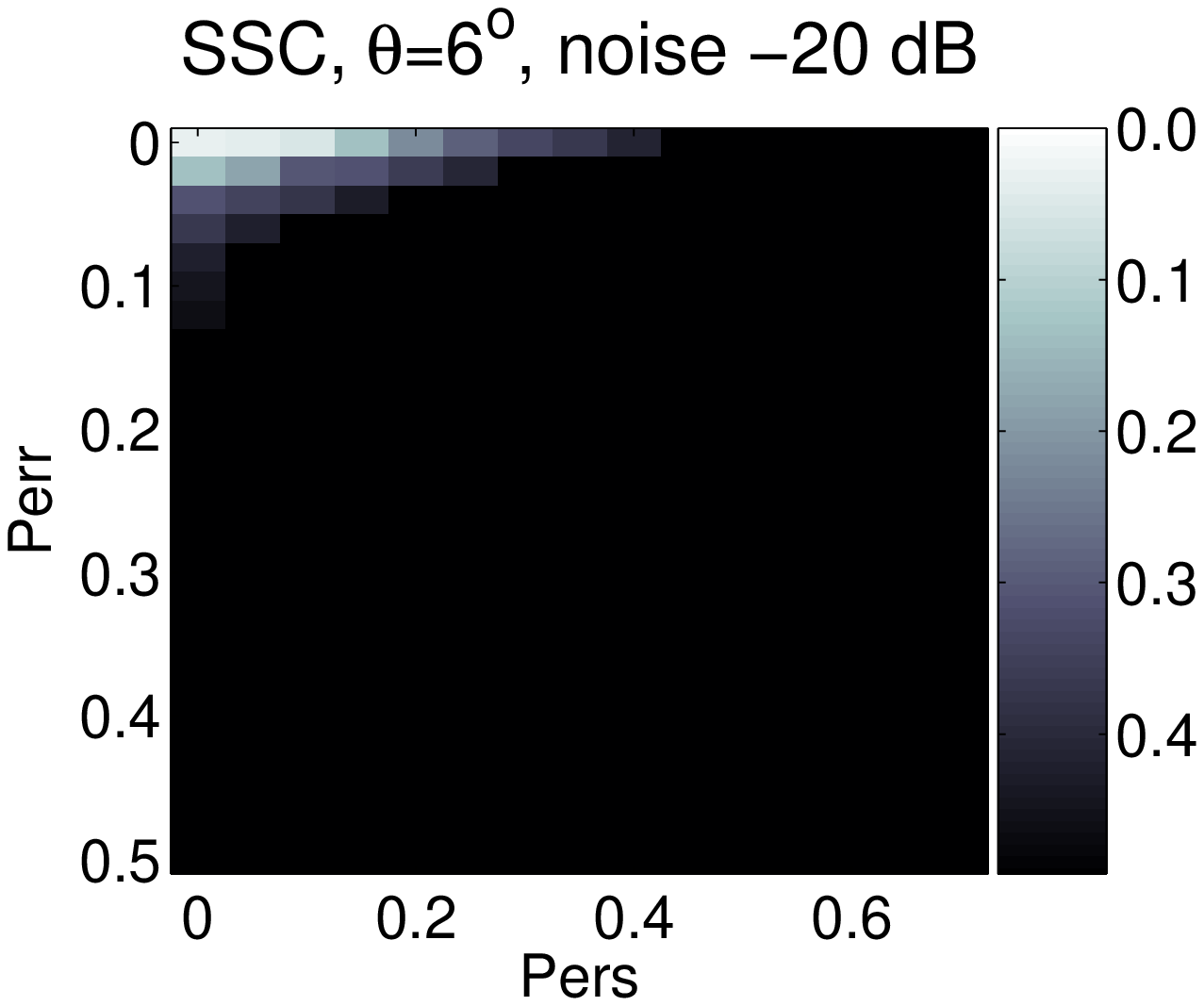,
           height=100pt, width=130pt}
		  } 
\put(120,5){          \epsfig{file=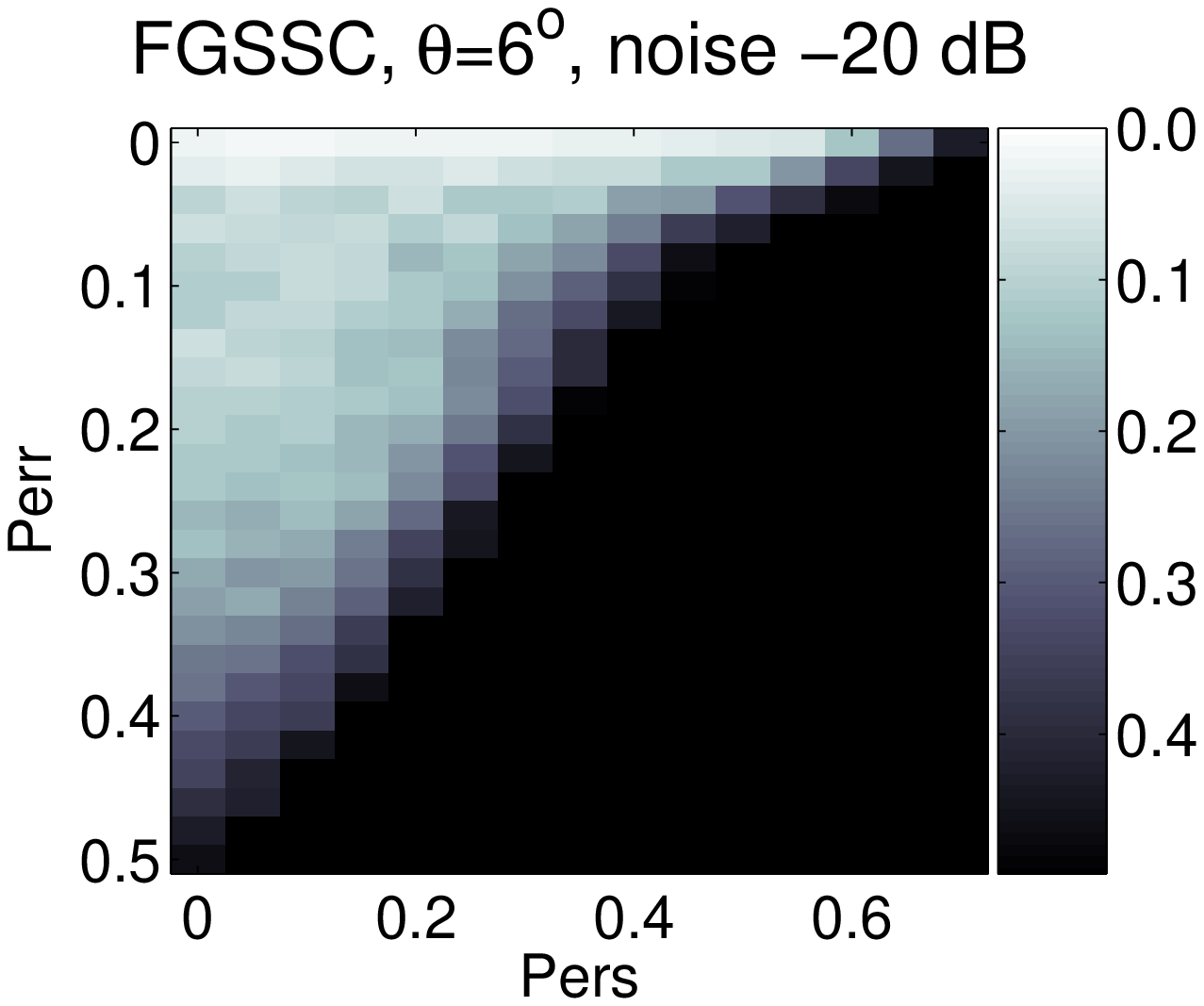,
           height=100pt, width=130pt}
		  } 

\end{picture}
\par
\noindent
\begin{picture}(0,115)
\put(0,10){          \epsfig{file=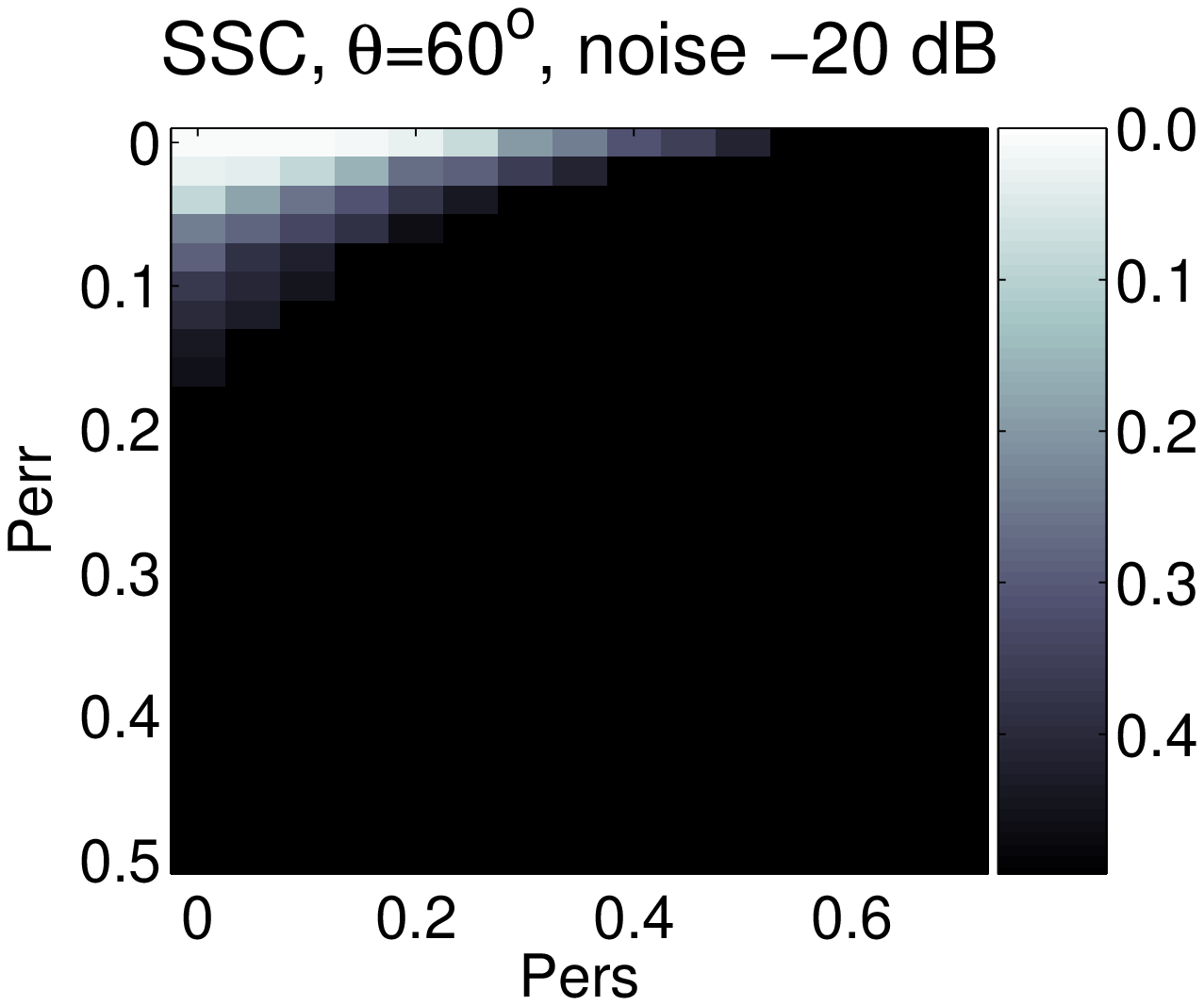,
           height=100pt, width=130pt}
		  } 
\put(120,10){          \epsfig{file=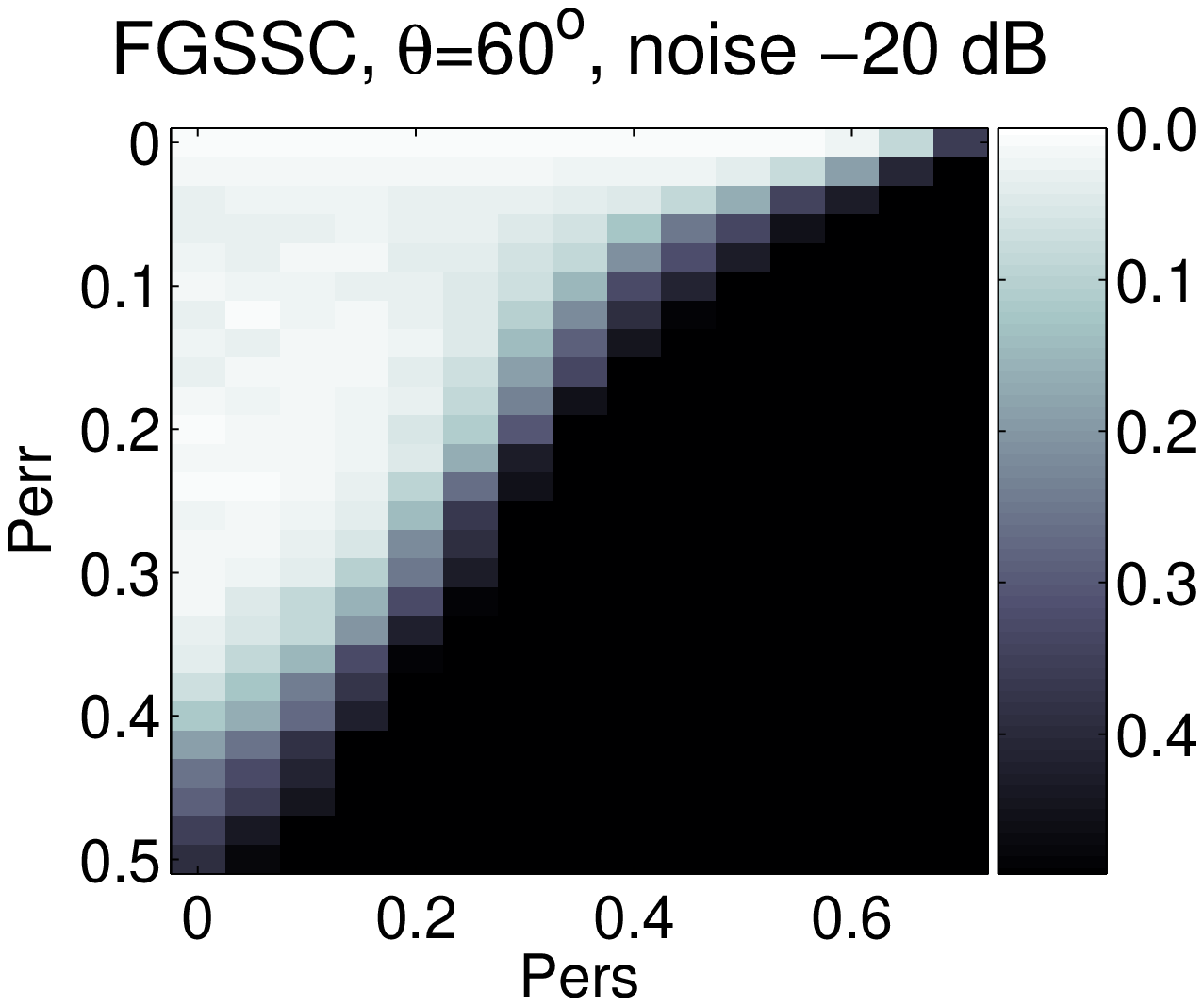,
           height=100pt, width=130pt}
		  } 
\put(50, 0){\footnotesize{\bf Fig.2. Clustering on noisy input, SNR=20 dB.} }
\end{picture}

\par
\noindent\par
We emphasize that all results on Figs. 1--3 were obtained  with the same algorithm settings. 
\par\noindent
\begin{picture}(00,115)
\put(0,10){          \epsfig{file=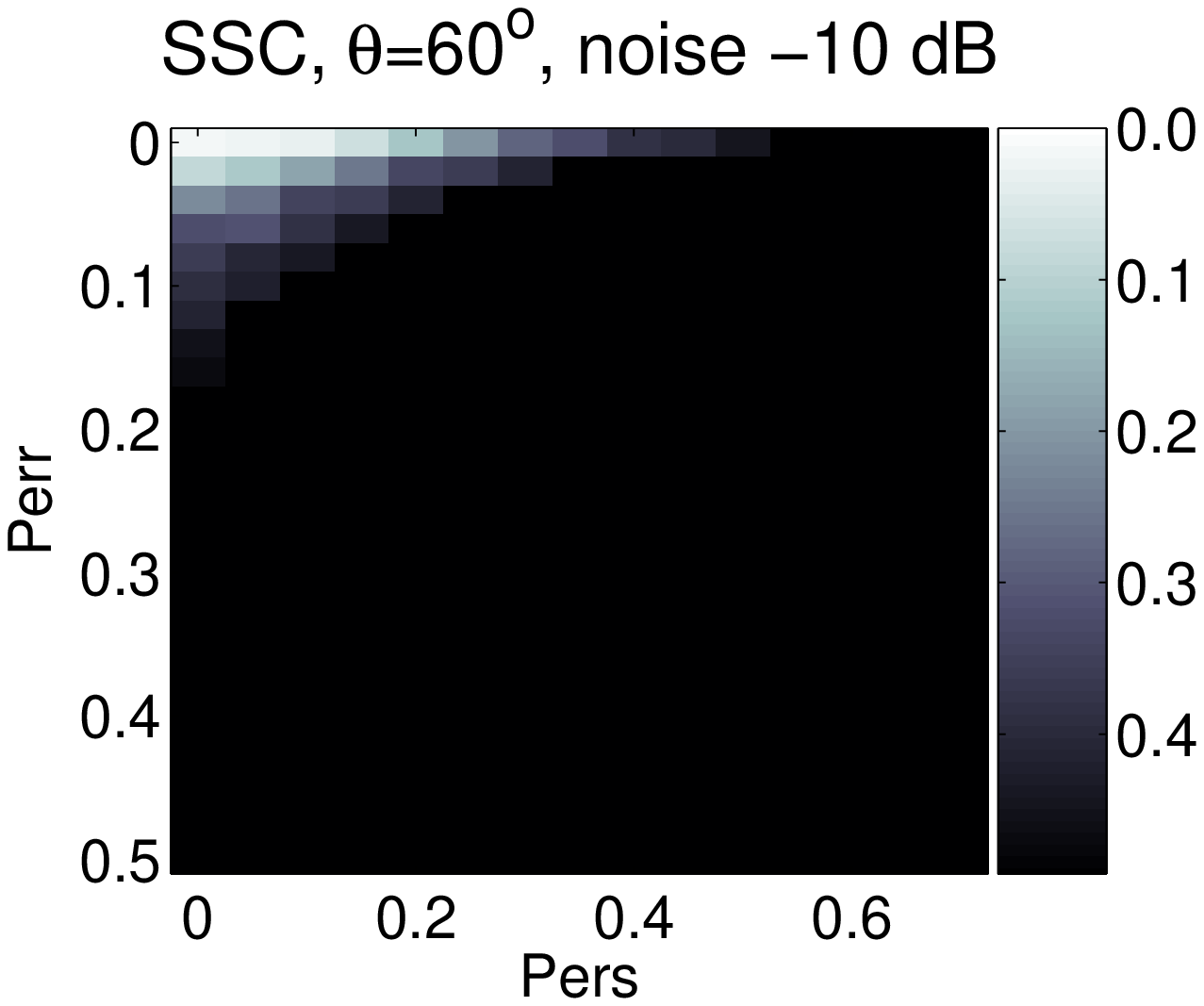,
           height=100pt, width=130pt}
		  } 
\put(120,10){          \epsfig{file=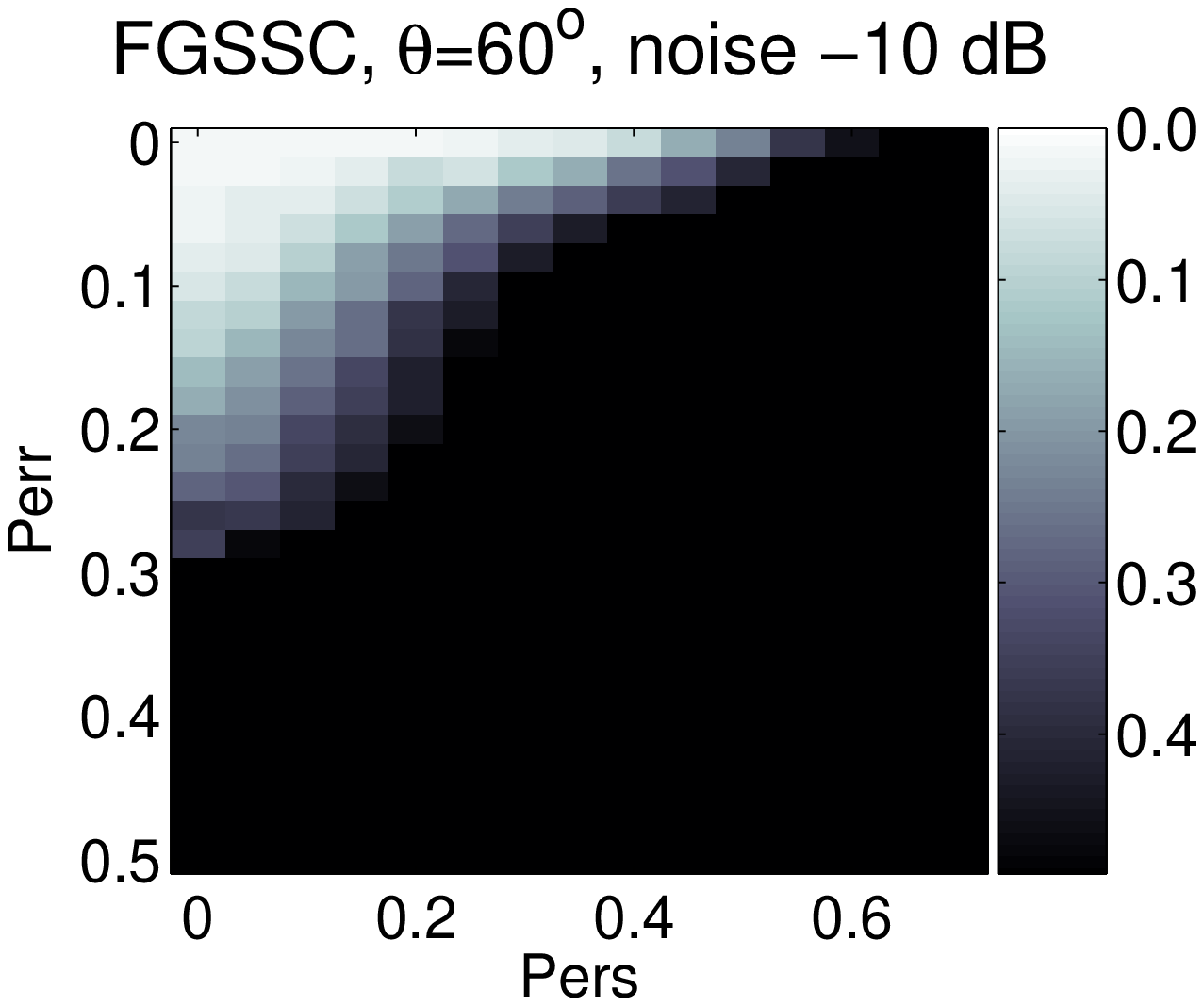,
           height=100pt, width=130pt}
		  } 
\put(50, 0){\footnotesize{\bf Fig.3. Clustering on noisy input, SNR=10 dB.} }
\end{picture}
\par\noindent

\par

\subsection{Synthetic Input II} 
We put our data  into the ambient space of the same dimension $D=50$ as in Section~\ref{synth1}.
However, we now use the model with absolutely random selection of subspace orientations. 
We fix the number of subspaces $K=7$ but the dimensions of subspaces are selected randomly
within the range $3\div 10$. 
In particular, those settings mean that the expectation of subspace dimensions is 6.5. Therefore, the total average dimension (the sum of all dimensions) of all subspace is close to the dimension of the entire ambient space. The number of data points is set as $N=200$. Those points are randomly distributed between planes. The number of points in each subspace does not depend on their dimensions.  However, to define a cluster and avoid obvious outliers the number of points per each subspace has to exceed the dimension of that subspace.
\label{synth2}
\par
\noindent
\begin{picture}(00,115)
\put(0,10){          \epsfig{file=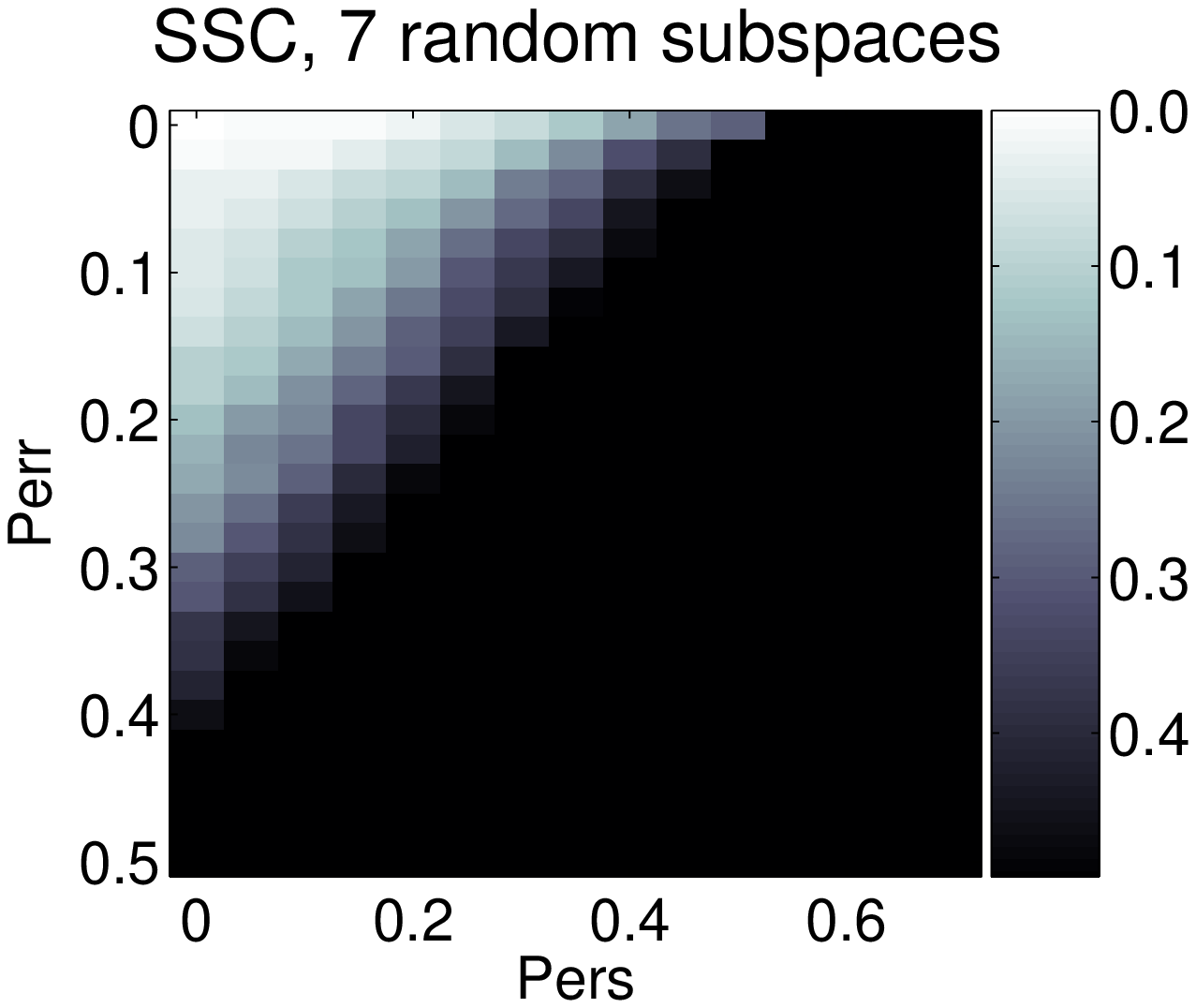,
           height=100pt, width=130pt}
		  } 
\put(120,10){          \epsfig{file=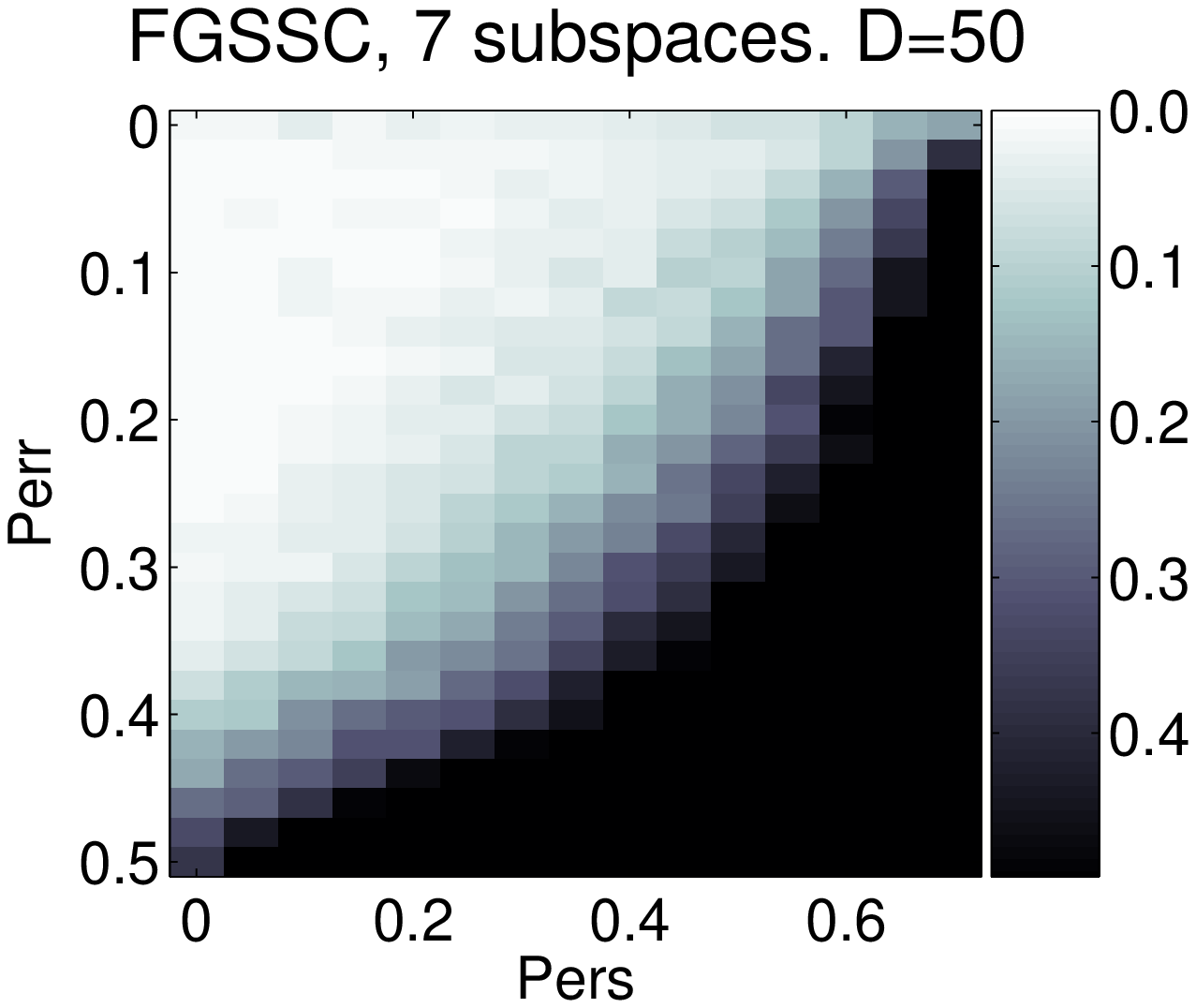,
           height=100pt, width=130pt}
		  } 
\put(70, 0){\footnotesize{\bf Fig.4. Clustering on 7 subspace model.} }
\end{picture}
\par
The parameter $\alpha_e$  has linear dependency from $P_{ers}$:
$$
\alpha_e:=\beta_1+\beta_2 P_{ers},
$$
whereas the dependence of $\alpha_z$ is quadratic:
$$
\alpha_z=\gamma_1+\gamma_2P_{ers}^2.
$$
The parameters are given in the Table 1. The parameters as well as the formulas were found empirically.
\par
\begin{table}[h]
\caption{Algorithm Parameters} 
\centering 
\begin{tabular}{cccc} 
\hline

    &$D=15$&$D=25$&$D=50$\\
\hline 
$\beta_1$& 5&5&7\\
$\beta_2$ &36 &22  &18\\
\hline 
$\gamma_1$ & 0.7 & 0.8 & 3\\
$\gamma_2$ &29 &10.8  &6.8\\
\hline 
\end{tabular}
\end{table}
\par
Now we present  results of two experiments with the lower dimension of the ambient space. Fig.~4 gives indirect hint about approximate possible dimension reduction. Indeed,  on Fig. 4, we see that  not perfect but decent FGSSC clustering is possible when only 30\% of data entries are available.
Erasing 70\% of the data entries in a vector can be interpreted as an orthogonal projection on the space  having lower dimension. In fact, introducing random erasures we just project our data on random coordinate planes.  If, instead of random erasures, we erase the  70\% of each data vector entries with the greatest indexes, we also have an orthogonal projection of all vectors on the same linear space.  While this procedure defines deterministic erasure which could lead to "systematic" drawbacks for some specific input, in our data model, the randomness is guaranteed by the input data. This non-rigorous common sense reasoning allows to expect that the reduction of the ambient space dimension at least up to $15=50\times 0.3$ is "equivalent" to 70\% erasures for the data in $\mathbb R^{50}$. The result of the~experiments for $D=15$ and $D=25$ are given on Fig.~5 and Fig.~6. They support our reasoning above. 

\par
\noindent
\begin{picture}(00,115)
\put(0,10){          \epsfig{file=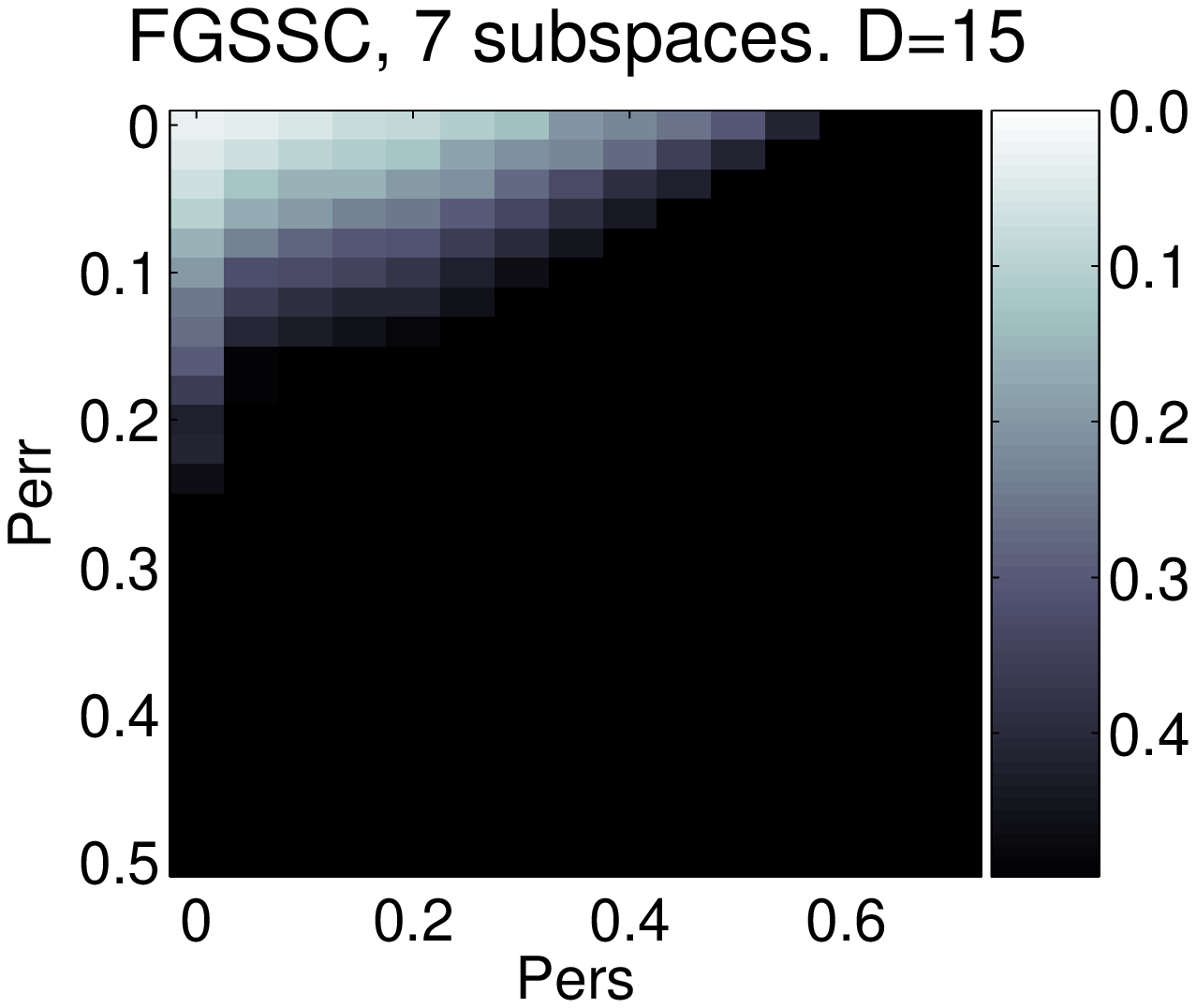,
           height=100pt, width=130pt}
		  } 
\put(120,10){          \epsfig{file=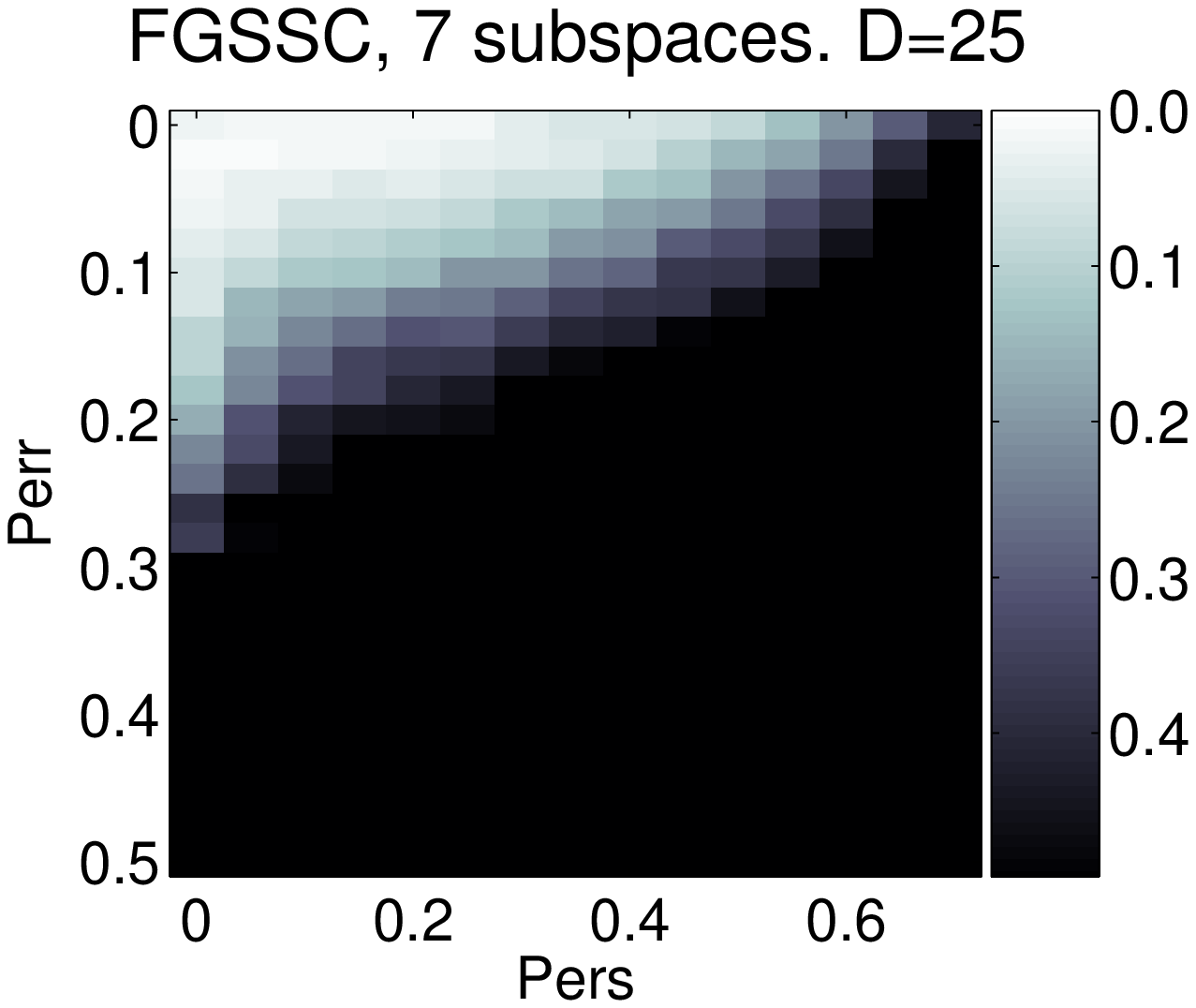,
           height=100pt, width=130pt}
		  } 
\put(0, 0){\footnotesize{\quad\bf Fig.5. Clustering for $D=15$.\qquad  Fig. 6. Clustering for $D=25$.} }
\end{picture}
\par
Actually, when  $D=15$ the efficiency of the algorithm for error and erasure free case is even higher than expected. 
We guess that in most cases the theoretical probability of correct clustering of a random dataset  with random locations of available entries  is equal to (or very close) to erasure free representation 
in the ambient space of the same number of available points. The similar result for suboptimal solutions associated with FGSSC algorithms could be extremely important for its theoretical justification. Indeed,
the original form of SSC from \cite{EV0} obtained detail study in \cite{SC} and \cite{SEC}.  The mentioned equivalence of the vectorwise and global projection would mean automatic transition
of all results from \cite{SC} and \cite{SEC} to the case of subspace clustering from incomplete data.
\par
We see that for the case $D=25$ and especially for $D=15$ the mean value of sum of dimensions of subspaces $\cS^i$ is a few times greater than the dimension of the ambient space. Say  $E[\sum_{i=1}^7 \dim\cS^i]=7\cdot6.5=45.5>3\cdot15=45$. Therefore, the sum of the dimensions is  more than 3 times greater than $D=15$. The input data are close to limits of algorithm applicability. Anyway, the ideas of Sparse Subspace Clustering are  still quite reliable even in such difficult case.
\subsection{Face Recognition}
\label{SectionFace}
 It was shown in \cite{BJ} that the set of all Lambertian reflectance functions (the mapping from surface normals to intensities) obtained with arbitrary distant light sources lies close to a 9D linear subspace. One of possible ways to use this result in combination with subspace clustering algorithms is the problem of face clustering. 
\par
Provided that face images of multiple subjects are acquired with a fixed pose and varying light conditions, the problem of sorting images according to their subjects is the obvious object for
trying the designed FGSSC. To our knowledge the state-of-the-art benchmarks are reached in \cite{EV}. Therefore, we organize our numerical experiments according to settings accepted in \cite{EV}. 
\par
We try the algorithms on the Extended Yale B dataset \cite{LHK}. The images of that dataset have
resolution $192\times 168$ pixels. We downsample those images to the resolution $48\times42$ by simple subsampling. Then we normalize their $\ell^2$-norm. Thus, each image is represented by a vector 
of dimension $D=2016$ with the unit length.
The data set  consists of images corresponding to 38 individuals. Frontal face images of each individual are acquired under 64 different lighting conditions.
We split those images into 4 groups corresponding to 1--10, 11--20, 21--30, 31--38 individuals.
We conduct our experiments inside those groups an then collect those results into total 
estimates for all groups. We estimate the efficiency of our algorithm on all possibles subgroups of $K=2,3,5,8,10$ individuals. For example, when we consider the case $K=3$ we conduct clustering experiments for $3 \binom{10}{3}+\binom{8}{3}=416$ triplets. Whereas, for $K=10$, since the 4th group contains 8 individuals, only 3 experiments are conducted.
\par
While, varying algorithm parameters for different input data (say for different $K$) the results can be significantly improved, 
in this section, we set the fixed parameters in all our trials. The adaptive selection of parameters by means of analyzing input data and intermediate results deserves a separate serious study.
\par
We use the following set of parameters for FGSSC.  $\epsilon=10^{-3}$, $\alpha=9.7$, $\alpha_z=81$, $\alpha_0=0.6$, $\alpha_1=1.0$, $\rho_0=1$, $\mu=1.02$.
\par
The results of processing are presented in Table 2.
\begin{table}[h]
\caption{Misclassification Rate (\%)} 
\centering 
\begin{tabular}{ccccccc} 
\hline

  &1 &2&3&4& FGSSC& SSC \cite{EV}\\
\hline 
{\it 2 subjects}\\
Mean &0.087 &0.071 &0.122&0.140 & 0.098 & 1.86\\
Median &0.00 &0.00  &0.00&0.00&0.00&0.00\\
\hline 
{\it 3 subjects}\\
Mean &0.234 &0.142 &0.191&0.995 & 0.297 & 3.10\\
Median &0.00 &0.00  &0.00&0.00&0.00&1.04\\
\hline 
{\it 5 subjects}\\
Mean &0.642 &0.248 &0.270&4.840 & 0.694 & 4.31\\
Median &0.00 &0.00  &0.00&0.31&0.00&2.50\\
\hline 
{\it 8 subjects}\\
Mean &1.280 &0.460 &0.768&16.2 & 0.949 & 5.85\\
Median &1.17 &0.40  &0.39&n/a&0.40&4.49\\
\hline 
{\it 10 subjects}\\
Mean &1.56 &0.64 &0.31&n/a & 0.84 & 10.94\\
Median &n/a &n/a  &n/a&n/a&0.64&5.63\\
\end{tabular}
\end{table}
The first 4 columns of Table 2 contain the average rate of misclassification in persents for FGSSC applied on groupwise data. Whereas the 5th column gives the FGSSC algorithm processing  results for consolidated data from all groups. The 6th column provides the results for the SSC algorithm reported in \cite{EV}.  Comparison of the results in columns 5 and 6 of Table 2 shows that FGSSC outperforms
SSC in  6--20 times in average misclassification.
\par
The low or zero values of the median of misclassification rate tells us that general principles of the algorithm are very reliable. They work perfectly on typical input data. The losses occur in some specific cases. The 4th group consisting of 8 individuals is an obvious outlier actively "spoiling" the higher results from other groups. To demonstrate how damaging outliers in the 4th group are, we give one example. One of triplet in the 4th group has misclassification rate 39\%. If only this triplet will be
clustered successfully like the overwhelming majority of other triplets of this group, the average misclassification over all triplets  in this group will drop   from 0.995 to 0.299. The thorough study of those exceptional cases may decrease the average misclassification value significantly. This is one of potential reserves of the greedy strategy. We may observe that the effect of low median is less evident for the  SSC algorithm.
\par
Another obvious way to improve the face recognition  is to incorporate the knowledge of the dimension (which is equal to 9) directly into clustering procedure. This option was not studied in this paper.
\par
The obtained results show that the idea of sparse self-representation utilize in the algorithm  is very deep and admits a lot of ways  for the efficiency improvement. Our greedy modification gave a significant decrease of misclassification rate. It is important to emphasize here that the algorithm is not aware
about any theory behind face recognition problem. It utilizes only general principles of linear algebra.

\section{Conclusions and Future Work}
We presented the Fast Greedy Sparse Subspace Clustering algorithm which is    a modification of the SSC algorithm based on a greedy approach. FGSSC has significant increased  the  resilience to
 corruption of entries on sparse set, data incompleteness, and noise. On the real database of images it provides 6--20 times lower rate of misclassification in face recognition than the SSC algorithm. It also significantly outperforms SSC on  a few models of synthetic data.
\par
Because of  very high capability of the algorithm, we believe that its theoretical justification as well as its
practical improvement is very desirable. We will mention a few possible directions of such development not addressed or studied not deeply enough in this paper.
\par
First of all we did not address any topics related to clustering algorithm itself.
Among crucial topics we would mention 3 most important of them. 
The~first one is finding outliers. The matrix $C$ has a lot of information about outliers because they do not fit the  property of subspace self-representation. Their occurrence can be discovered either from 
the~large $\ell^1$-norm of the corresponding column of the matrix $C$ or from too few entries in the 
final (after processing) matrix $\Omega$. The second topic is accurate estimation of the number of subspaces. The third one is a soft clustering decision allowing  to assign a few possible subspaces for points close to subspace intersections.
 \par
Our other suggestions for the future research are related to finding sparse representations, i.e., the matrix $C$.
\par
The original SSC algorithm as well as the FGSSC are based on finding a stationary point of functional (\ref{lagr}). They have quite strong resilience to data corruption, including  errors, erasures and noise. They provide high quality of clustering. However,  strictly speaking, they do not have  error correction capabilities. This is due to the fact that  in formula (\ref{lagr}) neither $E$ represents the values of errors nor $Y-YA-E$ represents  noise. 
\par
We believe that adding the error correction capability   may not only improve the clustering quality but also  have independent importance from the point of view of  processing of data located on several subspaces. This direction deserves the further research. Unfortunately, straightforward conversion of FGSSC for solving error correction problem would have too high computational complexity.
\par
One more reserve for algorithm improvement is selection of the parameters adaptive to input data.
The adaptation may bring significant increase of algorithm capability. One of such adaptive solution for error correction in Compressed  Sensing was recently found by the authors in~\cite{PK3}. 
\ifCLASSOPTIONcaptionsoff
  \newpage
\fi



%

%

\begin{IEEEbiography}{Alexander Petukhov}
received his Master (1982) and PhD (1988) degrees in
Mathematics from Moscow State University, Doctor (Habil.) of Sciences degree (1995) in Mathematics from Steklov Mathematical Institute (St.Petersburg). He became a Professor of Department of Mathematics of St.Petersburg Technical University in 1997. Since 2003, he is a faculty member of the Department of Mathematics of the University of Georgia.
\par
Alexander Petukhov is an expert in approximation theory and data representation. The theory of wavelet bases and frames, sparse data representations, signal, video and image representation and processing, digital film restoration are among his specific areas of interest.
\end{IEEEbiography}
\begin{IEEEbiography}{Inna Kozlov}
has a Ph.D in Mathematics from the Technion, Israel (1999), resolved difficult problems of Interpolation theory  of functional spaces (advanced integral norms); has an in-depth expertise and years of experience in acoustic signal processing, and successfully completed a number of industry projects which required ingenious handling of acoustics/seismic signals, development of advanced mathematical methods, and construction of efficient algorithms in signal and image processing. She headed Signal and Image Processing Division of a technology company Electro-Optics R\&D (Israel), working on a variety of commercial and military projects, and was the Head of Department of Computer Scince  in Holon Institute of Technology (Israel).  She is a founder and CTO of the high-tech company AlgoSoft Tech USA which is specializing on algorithms and software for Digital Film Restoration and Video Enhancement. 
 
Areas of research:
Wavelet analysis and Applications, Approximation Theory, Interpolation theory, Harmonic Analysis, Signal and Image Processing, Pattern Recognition, Compressed Sensing 
\end{IEEEbiography}
\end{document}